\documentclass[11pt]{article}
\PassOptionsToPackage{table}{xcolor}
\usepackage{template}
\usepackage{marvosym}
\usepackage{mathpazo}

\usepackage{fancyhdr}
\usepackage[T1]{fontenc}
\usepackage{ltablex}
\keepXColumns

\usepackage{microtype}
\usepackage[most]{tcolorbox}
\usepackage{pifont}
\usepackage{fontawesome5, wasysym, textcomp}

\hypersetup{
    colorlinks=true,
    filecolor=magenta,
    urlcolor=black,
    pdftitle={PaperDoctor},
    pdfpagemode=FullScreen,
}
\definecolor{citecolor}{HTML}{0071bc}
\hypersetup{citecolor=citecolor}
\definecolor{myorange}{RGB}{252,129,59}

\newcommand{\method}{PaperDoctor}
\newcommand{\eg}{\textit{e.g.,}}
\newcommand{\ie}{\textit{i.e.,}}

\definecolor{googlered}{HTML}{EA4335}

\definecolor{okgreen}{HTML}{2E8B57}
\definecolor{nored}{HTML}{C0504D}
\definecolor{midgray}{HTML}{9A9A9A}
\definecolor{lightgreen}{RGB}{240, 251, 237}
\definecolor{grouprow}{HTML}{EDEDED}
\definecolor{ourrow}{HTML}{FFF4D6}
\newcommand{\yes}{\textcolor{okgreen}{\ding{51}}}
\newcommand{\no}{\textcolor{nored}{\ding{55}}}
\newcommand{\yespart}{\textcolor{midgray}{\ding{51}}}

\newtcolorbox{methodbox}[2]{%
    enhanced,
    colback=#1!5,
    colframe=#1!40!black,
    boxrule=0.6pt,
    arc=2mm,
    left=2mm, right=2mm, top=3mm, bottom=2mm,
    title=#2,
    fonttitle=\bfseries\small,
    coltitle=white,
    colbacktitle=#1!40!black,
    attach boxed title to top left={xshift=4mm, yshift=-3mm},
    boxed title style={arc=1mm, boxrule=0pt, top=1pt, bottom=1pt, left=3pt, right=3pt}
}

\definecolor{wherecol}{HTML}{3B82F6}  
\definecolor{whycol}{HTML}{F59E0B}    
\definecolor{howcol}{HTML}{10B981}    

\title{\raisebox{-0.37\height}{\includegraphics[height=3.5em]{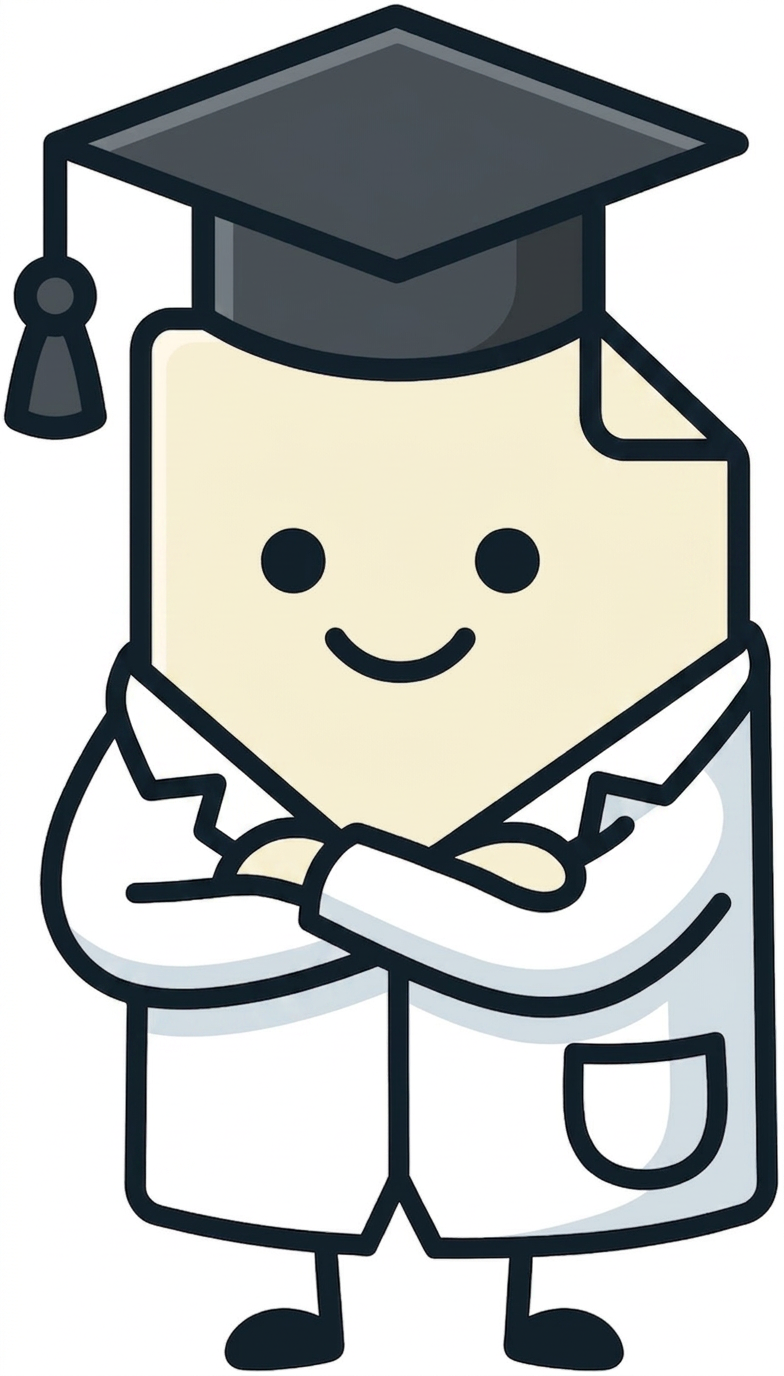}}\, {PaperDoctor: Evidence-Grounded and Actionable Feedback for Scientific Papers in Progress}}

\author[1]{Kevin Qinghong Lin}
\author[3]{Siyuan Hu}
\author[2]{Pan Lu}
\author[1]{Yu Chen}
\author[3]{Yanzhe Chen}
\author[2]{Owen Queen}
\author[1]{Yupeng Chen}
\author[1]{Jialin Yu}
\author[1]{Junchi Yu}
\author[5]{Zifeng Ding}
\author[2]{Yuanfeng Ji}
\author[2]{Sheng Liu}
\author[1]{Jindong Gu}
\author[4]{Linjie Li}
\author[3]{Mike Zheng Shou}
\author[1]{Philip Torr\textsuperscript{\Letter}}
\author[2]{James Zou\textsuperscript{\Letter}}
\vspace{-8mm}
\affil[1]{University of Oxford}
\affil[2]{Stanford University}
\affil[3]{National University of Singapore}
\affil[4]{University of Washington}
\affil[5]{University of Cambridge}

\date{}
\begin{document}
\maketitle
\vspace{-8mm}

\begin{center}
\href{https://github.com/QinghongLin/paperdoctor}{%
  \raisebox{-0.1\height}{\includegraphics[height=1em]{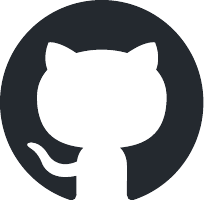}}%
  \hspace{0.3em}\textcolor{citecolor}{GitHub}
}%
  \quad \quad \quad
\href{http://paperdoctor.github.io/}{
  \raisebox{-0.25\height}
  {\includegraphics[height=1em]{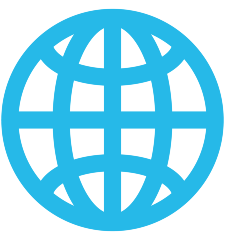}}
  \hspace{0.1em}\textcolor{citecolor}{Website}%
}
\end{center}

\begin{abstract}
{\normalsize
Autoresearch agents are reshaping the research ecosystem, but they can also let flawed claims enter the literature at scale. 
Human advisors catch such issues on in-progress drafts through careful, traceable feedback, yet advisor-style assessment requires extensive manual effort and does not scale.
To shift automated paper assessment from a judge to a diagnostician, we introduce \method, an agent framework for pre-submission feedback with three key innovations:
{(i) Holistic hierarchical framework.} Every paper is evaluated across writing, layout, references, code, theory, prior work, and experiments through three layers: 
L1 surface screening runs directly on every submission; 
L2 typed verifiers route each claim to the branch that validates the corresponding piece of evidence; 
and L3 reproducers rerun experiments by priority.
{(ii) Evidence-grounded actionable feedback.} Each \method's finding is a triple of an observation, a pointer to a specific evidence such as a sentence, equation, or code line, and a revision suggestion, making every critique auditable and actionable.
{(iii) Effective experimental reproduction.} Beyond reading the paper, \method\ selectively rebuilds and reruns experiments based on claim importance and compute budget, surfacing reproducibility gaps and quantitative limitations that are invisible from the manuscript alone.
We evaluate \method\ with 30 in-progress papers, yielding 70.6\% agreement and all positive holistic scores.
We further evaluate \method\ on 40 manuscripts across machine learning, natural and social-sciences, covering human- and AI-authored papers with code.
Overall, \method\ produces more auditable feedback than human and other agentic reviewers, pairs its critiques with concrete suggestions by design, and complements dimensions that are often overlooked by human reviewers.
To empower the community, we develop an interactive interface that lets authors browse findings grounded in their paper.
\method\ reframes automated paper assessment as a “\textit{diagnostic}” process rather than a verdict, taking a concrete step toward AI advisors that help with more rigorous AI-assisted scientific discovery.
\let\thefootnote\relax\footnotetext{{\Letter}: corresponding authors.}
}
\end{abstract}

\section{Introduction}
\label{sec:introduction}

\textit{``Don't find fault, find a remedy.''} --- Henry Ford

Autoresearch agents~\citep{aiscientist,aiscientistv2, agentlaboratory,airesearcher,agent4science} turn a proposed idea into a full manuscript, increasing the risk of producing work whose quality cannot be guaranteed, since such claims are often difficult to verify.
Moreover, most human-written drafts are now at least partly AI-assisted, whether in writing, figure drawing, or literature survey. This trend raises paper volume while leaving draft quality uncontrolled. 
Existing automated reviewers~\citep{marg,agentreview,mmreview, reviewer2,deepreview,remor} do not close this gap. By returning a decision (``accept'' or ``reject'') with a justification, they primarily serve to filter submissions, which is useful for the reviewer's side but uninformative for the author.

\begin{figure}[!h]
    \centering
    \includegraphics[width=\columnwidth]{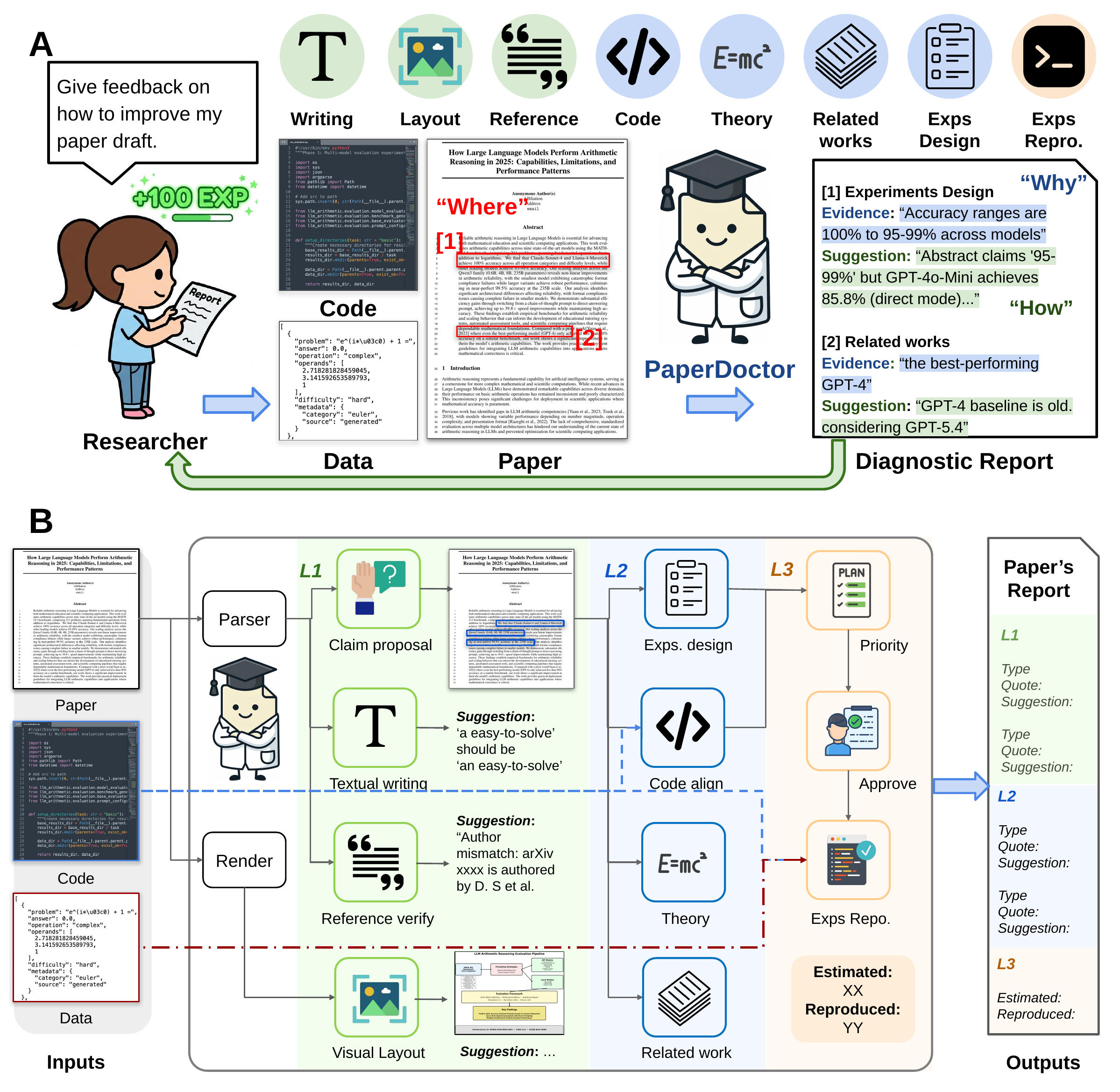}
\caption{\textbf{A. \method\ agent provides evidence-grounded, actionable feedback.}
Given a paper along with its code and datasets, \method\ returns a holistic report of findings across multiple dimensions (writing, citations, figures, equations, code, reproduction, and related work). Each finding pairs an error type (\textit{Why}) with a concrete suggestion (\textit{How}) and is grounded (\textit{Where}) in the paper itself, allowing authors to audit and learn from every critique.
\textbf{B. Illustration of \method\ pipeline.} 
PaperDoctor runs three diagnostic stages: L1 paper-only screening (Sec.~\ref{sec:method:l1}), L2 typed claim verifiers (Sec.~\ref{sec:method:l2}), and L3 prioritised experiment reproduction (Sec.~\ref{sec:method:l3}), to produce evidence-grounded, actionable feedback on pre-submission papers.
}
\label{fig:overall}
\end{figure}

Human advisors catch exactly these issues during discussion with junior researchers~\cite{can2011model, steiss2024comparing}. They go through the draft carefully: circling a claim with no experimental validation, underlining a theorem whose assumption breaks, or flagging a figure that is unclear. Each mark on the page is a small diagnosis. As illustrated in Figure~\ref{fig:overall} A, it points to {where} the symptom is in the paper, explains {why} it is a problem, and prescribes {how} to fix it; in short, it is \textit{evidence-grounded} and \textit{actionable}.
This is how researchers learn from an advisor's diagnosis on the page, which is fundamentally different from a reviewer's judgment. Yet such depth comes at a cost: it takes hours per paper, making it impossible to keep pace with agent-assisted writing~\cite{aiscientist,aiscientistv2,agentlaboratory,airesearcher}. 
Current agents either write the paper for the author~\cite{airesearcher,agent4science} or grade it at the door~\cite{liang2023llmreferee, lattereview,agentreview,reviewmt,reviewgraph}. Neither is what a researcher receives from an advisor: a careful diagnosis, like that of a doctor, that points to a specific section and says what to change.
However, delivering this level of feedback is non-trivial. A paper is a multi-faceted artifact spanning presentation, implementation, experimental analysis, and more. This is why authors typically rely on feedback from a range of people, such as advisors, peers, and industry mentors, each catching issues the others miss. 

Motivated by this, we ask: {can an agent provide diagnosis-level feedback at scale}?
We introduce \method, a research agent framework that delivers constructive feedback to human authors. 
Notably, \method~highlights the following:
\textbf{(i) Holistic, hierarchical pipeline.} \method\ decomposes a paper, together with its code and datasets, along dimensions ranging from writing to experimental reproduction, and organizes the assessment into three hierarchical levels of increasing cost and subjectivity. {L1 (surface screening)} operates directly on the manuscript itself at low cost, primarily targeting objective issues in writing, formatting, references, and layouts.
{L2 (claim verification)} extracts atomic claims and routes each one to the skill that owns its evidence—web search for prior-work checks, a vision language model (VLM) for figure assessment, code analysis for implementation claims, and theory verifiers for derivations. {L3 (experimental reproduction)} selectively executes experiments, validating reproducibility and correctness at execution time.
This design spends effort proportional to the cost of verification and keeps the full pipeline tractable.
{(ii) Evidence-grounded, actionable feedback.} Each finding is a triple of an observation with a reason (Why), a pointer to a specific location in the paper (Where, \eg~a sentence, equation, code line, or external URL), and a concrete suggestion for revision (How). This makes every critique both auditable and directly actionable for human authors.
{(iii) Effective experimental reproduction.} Beyond reading the manuscript, \method\ selectively prioritizes and reruns experiments based on claim importance and compute budget. By executing code rather than only reading it, \method\ surfaces reproducibility gaps and quantitative limitations that are invisible from the paper alone.

To validate the effectiveness of \method, we first conduct a human study with junior researcher participants (30 in-progress papers), collecting their in-progress paper drafts and asking them to rate the feedback produced by \method. \method\ reaches 70.6\% agreement with their evidence and all positive holistic scores.
Moreover, we evaluate \method\ on 40 manuscripts covering machine learning, the natural sciences, and the social sciences, including both human- and AI-authored papers with accompanying code and data.
This diverse benchmark allows us to systematically assess \method's robustness across disciplines, writing styles, and varying levels of methodological rigor.
We found that \method\ produces more auditable feedback than human and other agentic reviewers, pairs its critiques with concrete suggestions by design, and complements dimensions that are often overlooked by human reviewers. In particular, by executing the accompanying code and re-running key experiments, \method~reports where a published number could not be re-obtained under its protocol, a gap that reading-only review cannot see.

Lastly, to empower the research community, we release a demo interface (See Fig.~\ref{fig:demo}) that lets authors browse findings anchored directly on their own papers, making it easy to trace each critique back to its exact location.

\section{\method}
\label{sec:method}


\subsection{Overview}
\label{sec:method:overview}

Given a paper and its code, \method\ aims to produce a set of feedback $\{\mathcal{F}_i\}_{i=1}^{N}$, where each finding is defined as
\begin{equation}
  \mathcal{F}_i = \left(f_i,\, e_i,\, s_i\right).
  \label{eq:finding}
\end{equation}
Here, $f_i$ is the \textit{finding} (\ie~a brief textual description of the issue), $e_i$ is the \emph{evidence} grounding it to a specific location (a sentence, equation, code line, or external URL), and $s_i$ is a concrete \emph{suggestion} for revision. Notably, the pair $(e_i, s_i)$ lets the author audit and act on each finding without searching through the full paper.
Notably, we distinguish findings into two severity levels. An \emph{error} indicates that \method\ is confident the issue is incorrect (such as factual error), while a \emph{warning} indicates uncertainty and flags the finding for further clarification, such as a human check. 
We enable LLM to determine them autonomously, in order to better leverage context.

\paragraph{Paper-Code Parsing.}
Producing $\{\mathcal{F}_i\}$ by feeding the entire paper and codebase into a single model is infeasible: a paper is a long, multimodal document and a codebase is itself a large, structured artifact. Moreover, most downstream skills only need a targeted slice of the inputs (\eg~reference verification needs only the bibliography; figure assessment needs only the rendered pages). We therefore run a single parsing step that produces three decomposed reusable artifacts:
\begin{equation}
  \bigl(\mathcal{P}_t,\, \mathcal{P}_v,\, \mathcal{C}_t, \mathcal{B}\bigr)  \leftarrow (\texttt{Paper},\, \texttt{Code}),
\end{equation}
where $\mathcal{P}_t$ is the paper as section-organized markdown (via Mathpix\footnote{https://mathpix.com/}), $\mathcal{P}_v$ is the same paper rendered page-by-page as images for downstream Vision-Language Model (VLM) use, and $\mathcal{C}_t$ is the code indexed with tree-sitter\footnote{https://github.com/tree-sitter/tree-sitter} into per-file units. $\mathcal{B}$ denotes the parsed bibliography of the paper (\eg \texttt{.bib} file).
Every downstream skill reads from this shared representation and requests only the section, page image, or code snippet it needs.

\textbf{Hierarchical Pipeline.}
A paper spans many dimensions, and processing all of them in a single pass is infeasible. Different aspects demand different forms of evaluation, and these evaluations vary widely in cost: a writing check is a single LLM call, whereas experiment reproduction can consume hours of GPU execution.
We therefore organize \method\ as a hierarchical pipeline of three levels (L1--L3) that spends effort proportional to the cost of verification. As illustrated in Figure~\ref{fig:overall}B, L1 handles the most concrete, surface-level checks (such as presentation); L2 verifies individual claims along specific dimensions (such as theory or comparison with prior work); and L3 runs the most expensive stage, full experimental reproduction.

\subsection{L1 -- Surface Screening}
\label{sec:method:l1}
This stage focuses on straightforward issues that can be easily addressed by browsing the paper.

\textbf{Writing Review.}
We ask an LLM to read the paper by section $\mathcal{P}_t$ and flag writing issues as it goes. Clear mistakes such as typos or grammatical errors are marked as \textit{errors}, since they are unambiguously wrong. Stylistic issues, where the text is understandable but could be phrased more clearly, are marked as \textit{suggestions} instead, leaving the final decision to the author. For every issue, we require the LLM to quote the original sentence verbatim as evidence, ensuring each finding can be traced back to a specific location in the paper. 

\begin{methodbox}{green}{L1: Writing}
\small
\textbf{Evidence:} Page 3 \textit{``We trian the model on a large corpus of academic papers.''} \\
\textbf{Suggestion:} There is a typo: ``trian'' should be ``train''.
\end{methodbox}

\textbf{Figure Review.}
A paper is as much a visual artifact as a textual one: its figures, tables, and overall layout are carefully curated by the authors and judged by readers at a glance. Yet most of this visual information is lost in markdown extraction. We therefore treat visual inspection as a separate check in \method: we render the paper into page images $\mathcal{P}_v$ and ask a VLM to review them directly. Clear visual defects, such as figures overflowing the text margin or overlapping captions, are flagged as \textit{errors}. More subjective issues, such as undersized fonts or insufficient color contrast, are flagged as \textit{warnings} for the author to judge. 
As with writing review, every finding must be grounded to a specific page or figure index as evidence. 

\begin{methodbox}{green}{L1: Figure}
\small
\textbf{Evidence:} Page 5, Figure 3 extends beyond the right text margin. \\
\textbf{Suggestion:} Rescale the figure width to \texttt{\textbackslash linewidth}.
\end{methodbox}


\textbf{Citation Check.}
AI-assisted manuscripts routinely contain references that do not resolve to any real paper, and this is tedious to catch by reading the bibliography alone. Conditioned on $\mathcal{B}$, we pair the agent with a web search backend: for each reference, the LLM issues up to three search queries and records a resolver URL only when the backend returns a genuine match, never synthesizing itself.

\begin{methodbox}{green}{L1: Citation Check}
\small
\textbf{Evidence:} Reference [12] ``Smith et al., Neural Reasoning in Transformers, NeurIPS 2023'' returns no match. \\
\textbf{Suggestion:} The reference appears to be hallucinated. Please verify and replace with a valid source.
\end{methodbox}

\textbf{Claim Extraction.}
A paper makes dozens of arguments across its sections—novelty claims in the introduction, methodological choices in the method section, performance numbers in the experiments, and so on—and the value of \method\ comes from checking each of them against its own evidence. 
A claim that is never extracted can never be verified. We therefore ask an LLM to densely extract every verifiable assertion the authors make, covering \texttt{[theory, code, experiments(designs), literature]}; a single claim may be tagged with multiple evidence types. Each claim is then dispatched to the corresponding L2 branches for verification.

\begin{methodbox}{cyan}{L1: Claim Extraction}
\small
\textbf{Claim:} ``Our method achieves 92.3\% accuracy on ImageNet, outperforming prior state-of-the-art by 3.1 points.'' \\
\textbf{Evidence:} Introduction  \\
\textbf{Claim Type:} \texttt{[Experiment, Related Work]}
\end{methodbox}

Owing to the modular design, all four skills in L1 can run \emph{in parallel}.

\subsection{L2 -- Claim Verification}
\label{sec:method:l2}

In this stage, each claim extracted at L1 is sent to the corresponding verifier.

\textbf{Code Verification.}
A common failure mode of AI-assisted drafts is that the described optimizer, architecture, or training setup does not match the released code. These mismatches are almost invisible to human reviewers, who rarely open the repo while reading. We therefore ask \method\ to check each code-tagged claim directly against the source $(\mathcal{P}_t, \{\mathcal{C}_j\})$, which are indexable during our parsing stage.
For example, hyperparameter claims are often best verified by first inspecting configuration files before descending into the Python implementation: if the paper claims \texttt{AdamW} but the config specifies \texttt{Adam}, we flag a \textit{warning}---the mismatch is real, but may be a stale config or a last-minute switch the author should confirm. If a component described in the paper is missing from the code altogether, we flag it as an \textit{error}.

\begin{methodbox}{cyan}{L2: Code Verification}
\small
\textbf{Claim:} ``We train all models using the AdamW optimizer.'' \\[2pt]
\textbf{Evidence:} \texttt{configs/train.yaml} line 14 specifies \texttt{optimizer: Adam}, which conflicts with the paper’s experiment settings \\[2pt]
\textbf{Suggestion:} Mismatch between paper and code. Verify which optimizer was actually used.
\end{methodbox}

\textbf{Theory Verification.}
Errors in theoretical derivations are among the hardest to catch: a proof that reads smoothly often hides missing assumptions, skipped steps, or notation drift across equations, and even careful readers can miss these on a first pass. We therefore ask \method\ to re-derive each argument in $\mathcal{P}_t$ step by step rather than summarize it. \method\ jointly examines all theoretical content, including equations, variables, and the notation that links them across the paper, and records the full trace so that a superficial check is itself visible as a superficial trace. Specifically, \method\ examines four aspects in turn: correctness of each step, hidden assumptions that the paper does not state, boundary or edge-case behavior, and notation consistency across derivations. A loss whose expectation silently drops between its definition and its final form is caught here, before it propagates into code or experiments.

\begin{methodbox}{cyan}{L2: Theory Verification}
\small
\textbf{Claim:} ``The expected loss reduces to $\mathbb{E}[\|x - \hat{x}\|^2]$ (Eq.~7).'' \\[2pt]
\textbf{Evidence:} Step from Eq.~6 to Eq.~7 drops the cross-term $\mathbb{E}[x^\top \hat{x}]$ without justification. \\[2pt]
\textbf{Suggestion:} Missing assumption that $x$ and $\hat{x}$ are uncorrelated. Please state explicitly or correct the derivation.
\end{methodbox}

\textbf{Literature Check.}
A common issue in scientific drafts is overstated novelty or weak engagement with the literature (\textit{``no prior work addresses X''} when an earlier paper already does), and this bias is easier to catch by searching than by reading. Based on $(\mathcal{P}_t, \mathcal{B})$, \method\ pairs an LLM with a web search backend to separate baseline comparisons, cited facts, and novelty assertions, so that each novelty assertion can be further labelled as $\texttt{novel}$, $\texttt{incremental}$, or $\texttt{prior\_art\_exists}$.

Note that this differs from the reference verifier in L1: the reference verifier only checks whether a cited paper exists and is correctly attributed, while this stage examines whether the surrounding literature \emph{content} actually supports the paper's novelty and positioning claims.

\begin{methodbox}{cyan}{L2: Literature Check}
\small
\textbf{Claim:} ``No prior work addresses multi-modal reasoning over long video sequences.'' \\[2pt]
\textbf{Evidence:} Web search returns Chen et al. (2023), ``LongVid-Reasoner'', which targets the same setting. \\[2pt]
\textbf{Suggestion:} Novelty overstated. Relabel as {prior\_art\_exists} and cite Chen et al.
\end{methodbox}

\textbf{Experiment Design.}
While some claims can be closed-loop verified against external sources (literature) or formal content (theory, code), most claims in a paper rest on \emph{experimental evidence} and crucially, whether the experiments themselves are well-designed determines whether the contribution is genuinely grounded. 
Experimental issues thus fall into two regimes: design mistakes (missing ablations, missing experiments for a stated contribution) that can be found from the paper alone, and reproduction mistakes that can only be found by running. 
This module handles the first, before any execution. The agent reviews whether each argument is supported by a corresponding experiment, along with fairness, ablation sufficiency, statistical rigour, baseline recency, and cherry-picking risk. 
It then emits a \texttt{reproduction-plan} entry per experiment listing the command, priority, feasibility, run mode (evaluation or training), and the numeric target lifted from the paper. 
No experiments run here; the plan is a declarative contract for L3.

\begin{methodbox}{cyan}{L2: Experiment Design}
\small
\textbf{Claim:} ``Our cross-modal attention module is the key component driving the gains over prior work.'' \\[2pt]
\textbf{Evidence:} Table~3 reports only the full method versus the baseline; no ablation removes the cross-modal attention module to isolate its contribution. \\[2pt]
\textbf{Suggestion:} Add an ablation that disables the cross-modal attention module and reports performance on the same benchmark. \\[2pt]
\textbf{Reproduction plan:} \texttt{bash eval/ablate\_attn.sh}, target $\Delta\!\geq\!1.0$ point drop, priority high.
\end{methodbox}

Notably, the L2 stage remains highly modular: all four verifiers, as well as the per-claim dispatch within each verifier, run \emph{in parallel}.

\subsection{L3 -- Experiments Reproduction}
\label{sec:method:l3}

After L1 and L2, \method\ has already covered most aspects that can be assessed from the paper's main body. The remaining, equally important stage is reproduction, which is challenging yet essential. Different papers come with very different experimental setups: some require only inference, others involve full training, and many depend on substantial resources such as GPU compute, datasets, and storage. We therefore design a separate L3 stage dedicated to reproduction.

\textbf{Priority Ordering.}
It is worth noting that not every claim deserves an equal degree of attention.
For example, an experiment that backs a main claim in the abstract carries far greater weight than a hyper-parameter sensitivity study, even though the latter may be cheaper to run.
We rank experiments by their importance to the paper's central contributions and by their expected \textit{feasibility check}. \method\ assigns each entry in the reproduction plan a priority label of \texttt{\{high,medium,low\}}. Under a compute budget, high-priority items run before medium and low, evaluations before training, and ready experiments before blocked ones.

\textbf{Manual Approval.}
Reproduction consumes real compute and storage, and executes code that may affect the environment. A mis-typed claim could silently trigger a multi-hour training run. \method\ therefore presents the L2 plan, annotated with estimated GPU hours, dataset size, and storage footprint, for the author to approve. Only then does an LLM-driven executor handle environment setup, dispatch, and log parsing.

\begin{methodbox}{orange}{L3: Reproduction}
\small
\textbf{Claim:} ``Our method achieves 78.4\% accuracy on MMLU.'' \\[2pt]
\textbf{Evidence:} Reproduced run yields 71.2\% (\texttt{logs/mmlu\_eval.log}), a 7.2-point gap. \\[2pt]
\textbf{Suggestion:} Discrepancy exceeds the 1--2\% tolerance. Flagged as \textit{error}; verify evaluation protocol or reported number.
\end{methodbox}

\textbf{Report Results.}
Each executed experiment will receive a decision by comparing the reproduced value against the paper's reported one.
Rather than imposing a fixed numeric threshold, \method\ judges the verdict in context: it considers the metric type, the typical variance reported in the paper, and the magnitude of the original gap, and decides whether the result counts as a pass, a warning (partial match), or an error (execution failure or numeric mismatch).
This gives the author a direct view of which claims hold up under execution and which diverge from what the paper reports. 
\section{Results}
\label{sec:experiments}

\begin{figure}[!h]
    \centering
    \includegraphics[width=\linewidth]{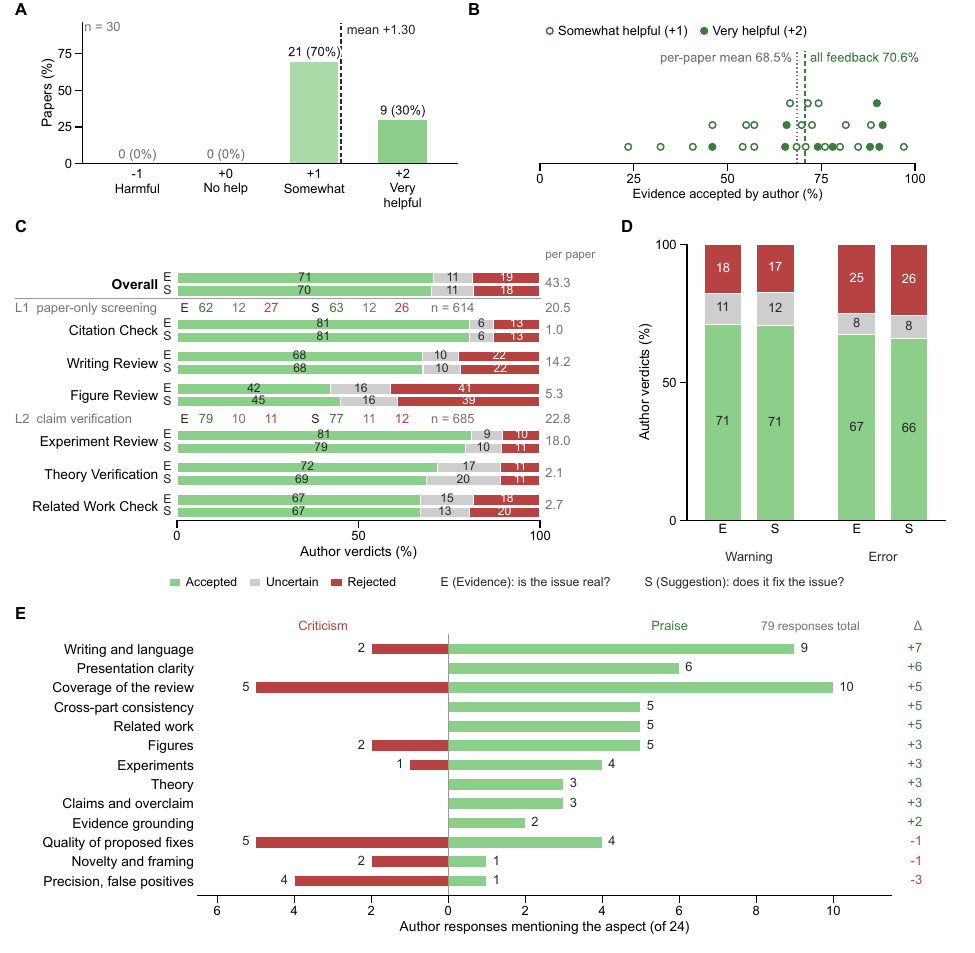}
    \caption{\textbf{Author verdicts on \method's feedback for 30 pre-submission papers, from the holistic review down to the fine-grained item.}
    \textbf{A.} Holistic rating of the feedback.
    \textbf{B.} Each paper's Evidence acceptance, by the rating its author gave the review.
    \textbf{C.} Author verdicts on Evidence (\textbf{E}) and Suggestion (\textbf{S}).
    \textbf{D.} Author verdicts within \method's own Warning and Error labels.
    \textbf{E.} The written comments that came with the ratings in \textbf{A}, by the aspect each addresses.
    }
    \label{fig:human-review}
\end{figure}

\noindent\textbf{Authors judge the review helpful overall, but agreement varies item by item}

\noindent Thirty pre-submission papers from 25 graduate students went through \method, and their authors judged every issue the six checks raised about the paper itself, 1,299 items in all.\footnote{We left out Code Verification. Pre-submission code is incomplete and uneven across papers, and Fig.~\ref{fig:repro} evaluates that stage separately.} Authors rated each item accepted, uncertain, or rejected on two axes: Evidence, whether the issue is real, and Suggestion, whether the proposed fix resolves it. They also scored the review as a whole from $-1$ (harmful) to $+2$ (very helpful), with an optional free-form comment. All 30 scores were positive, 70\% somewhat helpful and 30\% very helpful, for a mean of $+1.30$ (Fig.~\ref{fig:human-review}A).

Authors accepted most of what \method\ found evidence on their own paper (Fig.~\ref{fig:human-review}B). Across all 1,299 items the acceptance rate was 70.6\%, and the mean of the 30 per-paper rates was 68.5\%. Per paper the rate ran from 23.5\% to 97.0\% with a median of 71.1\%. The authors at both extremes also rated the review $+1$, and across the 30 papers the correlation between Evidence acceptance and the holistic score is positive but not statistically significant ($r = +0.29$, $P = 0.12$). The two measurements are not substitutes: the score asks whether the feedback improved the draft, and the item ratings ask whether each evidence is real.

Authors agreed more often with the L2 verifiable claims than with the L1 surface screening (Fig.~\ref{fig:human-review}C). The three L2 checks were rejected 11\% of the time, the three L1 checks 27\%. Experiment designs drew the most items of any claim, 539, and authors accepted 81\% of them. The Figure Review assesses how effectively each figure is designed, and authors accepted 42\%, less than for any other claim. Vision perception models still misread dense panels, and much of figure design is the author's own decision. Citation check raised the fewest items, because it confirmed most of the references it read rather than flagging them.

Authors were more certain about the items \method\ marked Error, and agreed with them less (Fig.~\ref{fig:human-review}D). They accepted 71\% of Warnings on both axes, against 67\% of Errors on Evidence and 66\% on Suggestion. Marking an item an Error cut the uncertain share from 11\% to 8\% and raised the rejected share from 18\% to 25\%. Accepting that an issue was real almost always meant accepting the suggestion proposed for it: of the items whose Evidence an author accepted, 97.1\% had the Suggestion accepted too, against 3.3\% of the items whose Evidence they rejected.

Authors split over how much the review should cover (Fig.~\ref{fig:human-review}E). An LLM split the 24 written comments into 79 aspect responses, 58 praise and 21 criticism. No aspect drew more praise than breadth, at 10 responses, and none drew more criticism, at 5. Writing and language drew the next most praise, 9 responses, and it is the part of a paper a language model is best placed to check. Presentation clarity followed at 6, then cross-part consistency, related work, and figures at 5 each, three aspects that each need evidence from outside the sentence in front of the reader.
Three aspects drew more criticism than praise: the fixes \method\ proposed, the items it got wrong, and its silence on novelty and framing. The fixes and the silence ask \method\ to judge what is worth saying rather than to check whether something is true. Scanning every dimension of every paper also costs precision: the items \method\ got wrong drew 4 criticisms against 1 praise, the widest negative margin in the panel.

\begin{figure}[!h]
  \centering
  \includegraphics[width=\linewidth]{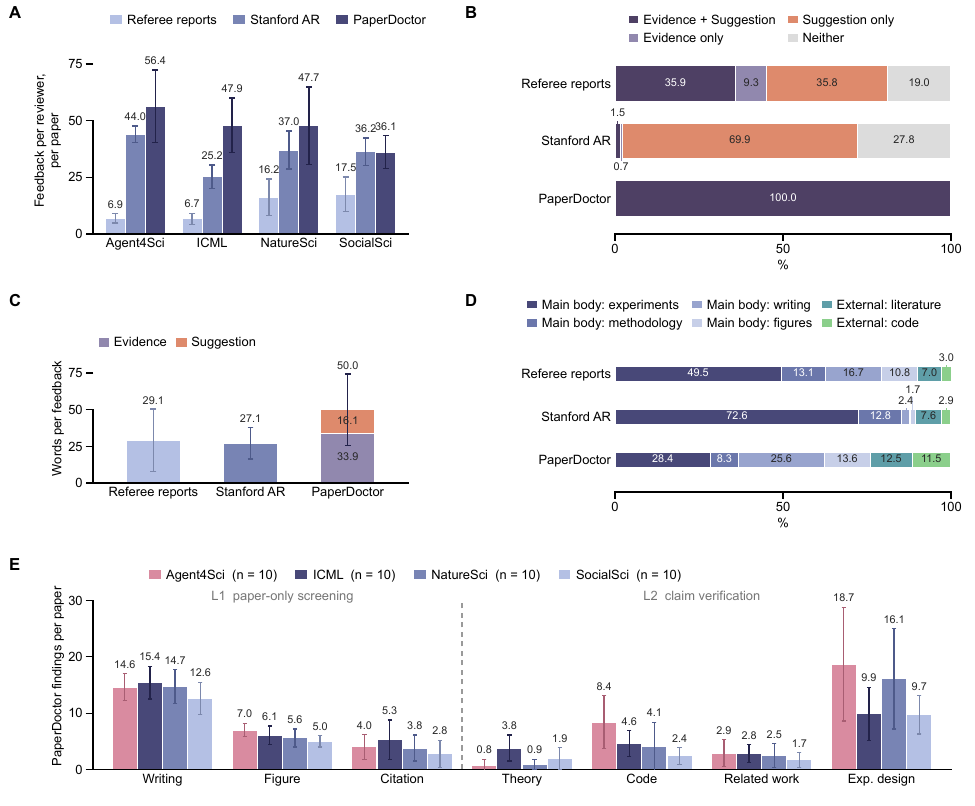}
  \caption{\textbf{Comparison between referee reports, the Stanford Agentic Reviewer and \method\ on the same 40 papers, from how much each writes down to what it is about.}
  \textbf{A.} Findings per reviewer, per paper, by domain.
  \textbf{B.} Whether a finding carries Evidence, a Suggestion, both, or neither.
  \textbf{C.} Words per piece of feedback, \method's split into its Evidence and its Suggestion.
  \textbf{D.} What the findings are about.
  \textbf{E.} \method's findings per paper, by check and by the kind of paper reviewed.
  Bar, the mean; whisker, one standard deviation. First published referee round only.
  }
  \label{fig:reviewer-comparison}
\end{figure}

\noindent\textbf{Evidence-grounded feedback arise by design, not by chance.}
\label{sec:exp:vs-reviewers}

\noindent We compared \method\ against the reviews the same papers actually received. The 40 papers are four groups of 10: Agents4Science~\cite{agent4science}, whose papers are AI-written and whose reviews are themselves AI-written; ICML orals and spotlights; Nature Communications, for the natural sciences; and Nature Human Behaviour, for the social sciences.
Every paper carries three feedback of itself: its published referee reports; one run of the Stanford Agentic Reviewer\footnote{\url{https://paperreview.ai/}, Stanford ML Group.}, an agentic system that reviews from the paper alone; and our \method\footnote{We count only the first published referee round, because neither automated system ever sees a revision}. 
Because the three sources have different output structures, we use Gemini-2.5-Flash to parse all feedback into a common set of atomic points for consistent analysis.

\method\ writes the most and with the widest spread, and the referee reports the least (Fig.~\ref{fig:reviewer-comparison}A). We compare per reviewer\footnote{The nature-series journal paper is read by 2.6 referees and an Agent4Sci paper by 3.2 while the two automated systems are one each}.
The Agent4Science and ICML conference reports contain 6.9 and 6.7 findings per source, respectively, compared with 16.2 and 17.5 for the Nature journal reports. This difference remains consistent across referees. Both automated systems write more than that in every domain. \method\ writes 36.1 to 56.4 findings per reviewer against the agentic reviewer's 25.2 to 44.0, ahead of it in three domains and level with it in the fourth. \method\ shows larger within-field variation, suggesting higher feedback sensitivity to the individual paper, with a range of 53 findings on Agent4Science papers versus 13 for the agentic reviewer.

\method\ differs from the agentic reviewer most of all in evidence grounding (Fig.~\ref{fig:reviewer-comparison}B). Both referees and the agentic reviewer propose a Suggestion at nearly the same rate (in 71.7\% and 71.5\% of their findings), showing that the agentic reviewer provides suggestions about as often as human referees.
When it comes to Evidence, which requires grounding in the paper, referees provide it in 45.2\% of their findings, compared with only 2.2\% for the agentic reviewer.
Considering Suggestion and Evidence jointly, 35.9\% of referee findings contain both, versus 1.5\% for the agentic reviewer; instead, the agentic reviewer’s most common pattern is a suggestion without supporting evidence, accounting for 69.9\% of its findings.
Every \method\ finding carries both, because the design explicitly design for this: \method\ never emits a record with no quote and no suggestion. 
It pays for that in length, averaging 50.0 words per finding against a referee's 29.1 and the agentic reviewer's 27.1, of which 33.9 are the Evidence and 16.1 the Suggestion (Fig.~\ref{fig:reviewer-comparison}C).

\method\ covers a paper more evenly than either reviewer, and looks outside it more often (Fig.~\ref{fig:reviewer-comparison}D). 
Both referee and agentic reviewer spend mostly on the paper main body, 90.1\% and 89.5\% of their findings, and the experiments draw most of that: 49.5\% of a referee's findings and 72.6\% of the agentic reviewer's. \method's distribution is the flattest of the three: the experiments are its largest single dimension, at 28.4\%. It is the source that spends much outside the paper, 12.5\% on the external literature and 11.5\% on the code against 10.0\% and 10.5\% for the two on both dimensions together. 

Code and Experiment design are the two checks that tell the groups apart (Fig.~\ref{fig:reviewer-comparison}E). The three L1 checks (writing, figure, and citation) report comparable numbers everywhere.  L2's Code separates the groups by authorship, 8.4 findings per paper for the AI-written Agent4Science papers against 4.6, 4.1 and 2.4 for the three human-written groups. 
Related work is flat at 1.7 to 2.9, and Theory rises only for ICML, at 3.8 against 0.8 to 1.9 elsewhere, where the papers are about machine-learning methodology. 
Experiment design draws the most findings of the four L2 checks and the widest spread with them, 18.7 for Agent4Science and 16.1 for Nature Science against 9.9 for ICML and 9.7 for SocialScience. Both L2 Code and Experiment Design suggest that reproducibility may be a key factor distinguishing the paper groups.

\begin{figure}[!t]
  \centering
  \includegraphics[width=\linewidth]{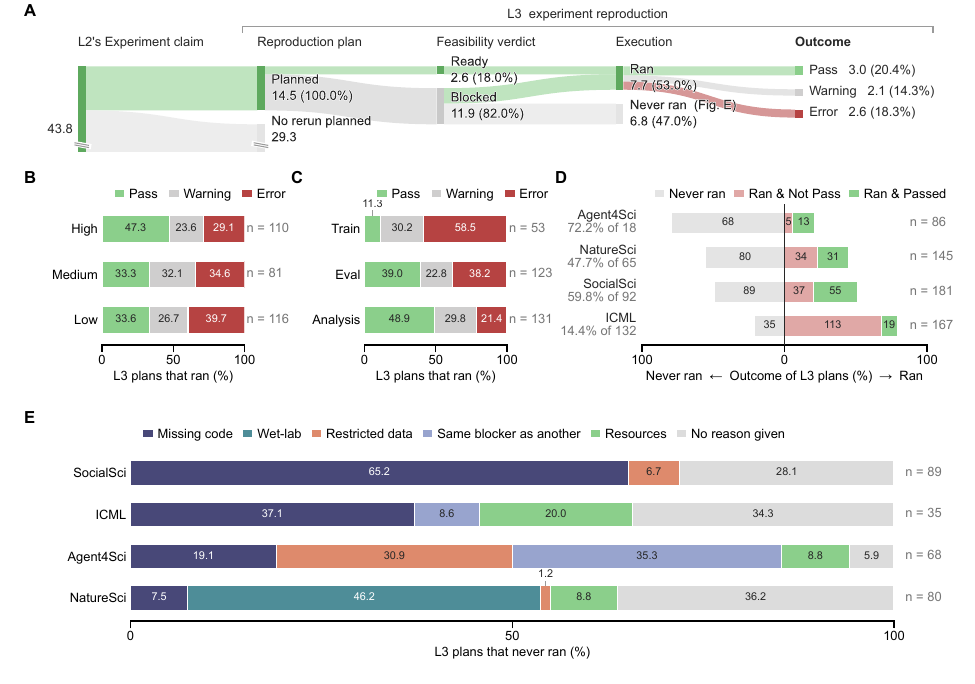}
  \caption{\textbf{Reproducing the experiments of 40 papers that release code and data.}
  \textbf{A.} From a claim to a number, per paper; green, still moving; grey, stopped.
  \textbf{B.} Outcome by the rank the plan carried, over the plans that ran.
  \textbf{C.} Outcome by what the plan reruns, over the plans that ran.
  \textbf{D.} What became of a source's plans; the centre line is a command running.
  \textbf{E.} Why the plans that never ran did not run, by source.
  Every node in \textbf{A} is a per-paper average, with its share of the reproduction plans beside it.
  A plan counts as run when a command was actually executed for it.
  }
  \label{fig:repro}
\end{figure}

\noindent\textbf{Reproduction is bottlenecked twice, before a command runs and after it}
\label{sec:exp:repro}

We sent the L2 Experiment claims from the same 40 papers to the reproduction stage. L3 turns claims that require re-execution into executable plans with verifiable targets, such as reported numbers. Each plan is then assigned a pass, a warning, or an error. Error groups execution failure and numeric mismatch.
Of the 43.8 experimental claims per paper, 14.5 become reproduction plans, but only 18.0\% of these plans are judged ready to run, while 82.0\% are initially blocked. After agent resolving feasible blockers, 7.7 plans reach execution, and 5.1 return a pass or warning (Fig.~\ref{fig:repro}A). 

Reproduction is limited by missing guideline and mismatched reproduced results.
Almost half of all plans, 47.0\%, never reach a command at all, most often because the paper's repository does not provide enough information to determine how to run them (Fig.~\ref{fig:repro}E). 
Even getting a command ready to run is still far from validating the claim: among the 53.0\% of plans that reach execution (Fig.~\ref{fig:repro}A, right), 27.0\% result in warnings and 34.5\% in errors.

A plan that \method\ ranked highly passes more often, and errors less often (Fig.~\ref{fig:repro}B). High-priority plans pass 47.3\% of the time once they run and error 29.1\% of the time. Low-priority plans pass 33.6\% and error 39.7\%. Medium sits with low rather than between the two, passing 33.3\%. \method's prioritization is consistent with human practices: important components are often prepared and maintained more carefully, making high-priority plans more likely to reproduce successfully.

A plan that requires training passes least often and errors most (Fig.~\ref{fig:repro}C). Training passes 11.3\% of the time once it runs and errors 58.5\%. Rerunning inference (such as based on released checkpoint) passes 39.0\% and errors 38.2\%, and rerunning statistical test or analysis pipeline passes 48.9\% and errors 21.4\%. 
The order follows how much of the original setup a run has to rebuild. Training needs the environment, the dataset and the compute budget together; a checkpoint needs the first two; an analysis pipeline simply needs the data alone.

The AI-written papers and the ICML oral/spotlight papers exhibit two contrasting failure modes (Fig.~\ref{fig:repro}D). 
ICML papers usually fail after a command runs: 132 of 167 plans execute, but only 19 match the reported number (14.4\%). Agent4Science papers usually fail before a command runs (68 of 86), yet 13 of the 18 that do run match (72.2\%). NatureScience and SocialScience sit between the two, losing about half at each of the two stages.

Different field papers exhibit distinct and representative causes of failure (Fig.~\ref{fig:repro}E). Incomplete code stops 65.2\% of SocialScience's blocked plans and 37.1\% of ICML's: the paper deposits its data, but not the analysis scripts that produced its paper figures. NatureScience is stopped mostly by environments the agent cannot rerun: 46.2\% of its blocked plans rest on wet-lab procedures its discipline cannot move to a machine. Agent4Science is stopped earlier still, by restricted data and by near-duplicate plans that carry another plan's blocker. 
Across all 272 blocked plans, an incomplete runnable environment (either missing code or model weights) is the largest cause, accounting for 33.1\%.
\begin{table*}[!t]
\centering
\small
\setlength{\tabcolsep}{3.5pt}
\caption{\textbf{\method\ vs.\ representative research agents}.
\textbf{Reviewer Report}: produces reviewer-style
prose, not only a score. \textbf{Grounded Evidence}: every finding is
anchored to a concrete span/figure/equation/code line.
\textbf{Revision Suggestion}: output specifies \emph{what to change}, not
only \emph{what is wrong}. \textbf{Text (LLM)}: reads body text,
tables, and document structure. \textbf{Visual (VLM)}: reads figures as
images. \textbf{Code Audit}: cross-checks paper claims against released
code. \textbf{Exp. Reproduction}: re-executes experiments.
\textbf{In-progress papers}: can the agent support or focus on these in-progress manuscripts.
\yes=full, \yespart=partial, \no=absent.}
\label{tab:related-compare}
\begin{tabular}{l|ccc|cccc|c}
\toprule
 & \multicolumn{3}{c|}{\textbf{Feedback Form}}
 & \multicolumn{4}{c|}{\textbf{Assessment Coverage}}
 & \textbf{Focus} \\
\cmidrule(lr){2-4} \cmidrule(lr){5-8} \cmidrule(lr){9-9}
\textbf{System}
 & Reviewer  & Grounded  & Revision
 & Text       & Visual  & Code    & Exp.
 & In-progress \\
 & Report    & Evidence  & Suggestion
 &  (LLM)  & (VLM)   & Imple.   & Repro.
 & Papers \\
\midrule
\rowcolor{grouprow}
\multicolumn{9}{l}{\textit{\textbf{Autoresearch}} 
} \\
AI Scientist~\citep{aiscientistv2}       & \yes & \no  & \no       & \yes & \no  & \yespart & \yes  & \no  \\
\midrule
\rowcolor{grouprow}
\multicolumn{9}{l}{\cellcolor{grouprow}\textit{\textbf{Peer Review}}} \\
MARG~\citep{marg}                        & \yes & \no  & \yespart  & \yes & \no  & \no      & \no  & \no  \\
AgentReview~\citep{agentreview}          & \yes & \no  & \yespart  & \yes & \no  & \no      & \no  & \no  \\
Reviewer2~\citep{reviewer2}              & \yes & \no  & \yespart  & \yes & \no  & \no      & \no  & \no  \\
DeepReview~\citep{deepreview}            & \yes & \no  & \yespart  & \yes & \no  & \no      & \no  & \no  \\
TreeReview~\citep{treereview}            & \yes & \yespart & \yespart & \yes & \no  & \no    & \no  & \no  \\
CycleResearcher~\citep{cyclereviewer}      & \yes & \no  & \no       & \yes & \no  & \no      & \no  & \no  \\
MMReview~\citep{mmreview}                & \yes & \yespart & \no   & \yes & \yes & \no      & \no  & \no  \\
\midrule
\rowcolor{grouprow}
\multicolumn{9}{l}{\textit{\textbf{Verification}}} \\
CiteAudit~\citep{citeaudit}              & \no  & \yes & \no       & \yes & \no  & \no      & \no  & \no \\
PaperBench~\citep{paperbench}            & \no  & \yespart & \no   & \yes & \no  & \yes     & \no & \no \\
AutoReproduce~\citep{autoreproduce}          & \yes & \yespart & \no & \no & \no & \no & \yes & \no\\
\midrule
\multicolumn{9}{l}{\cellcolor{grouprow}\textit{\textbf{Feedback}}} \\
Review feedback~\cite{reviewfeedback} & \yes & \yes & \no & \yes & \no & \no & \no & \no\\
\rowcolor{lightgreen}\textcolor{gray}{Human advisor}  & \yes & \yes & \yes      & \yes & \yes & \yespart     & \no & \yes \\
\rowcolor{lightgreen}
\textbf{\method\ (ours)}                 & \yes & \yes & \yes      & \yes & \yes & \yes     & \yes & \yes \\
\bottomrule
\end{tabular}%
\end{table*}

\section{Related Work}
\label{sec:related}

\paragraph{AI for Autoresearch}
End-to-end autoresearch pipelines chain ideation, experimentation, and drafting into a single agentic loop~\citep{aiscientist,aiscientistv2,agentlaboratory,airesearcher,agent4science}, increasingly supported by language and coding agents~\citep{researchcodebench,lmrbench,scireplicate} that probe whether agents can implement and run published methods. Recent variants explore tool augmentation, multi-agent collaboration, and long-horizon execution, so manuscripts are produced with ever less human oversight.
None of these pipelines, however, includes an internal quality-control stage, so characteristic failure modes pass silently into the final draft: hallucinated citations, inflated novelty, mismatches between claims and the code that implements them, and unverified empirical statements. \method\ is complementary to this line of work: rather than producing papers, it consumes a draft together with its accompanying code and data, decomposes it across writing, references, theory, and experiments, and locates the parts that need revision before submission. In this sense, it is the internal advisor that current autoresearch agents lack.

\paragraph{Paper Verification}
For rigorous assessment, a paper can be treated as a verifiable system whose claims must be checked along multiple dimensions.
(\romannumeral1) {Citation verification.} LLM drafts routinely contain fabricated references, and audits report elevated hallucination rates~\citep{ghostcite,citehallucinationstudy}; dedicated verifiers~\citep{citeaudit,wu2025automated} and attributed-generation frameworks~\citep{gao2023enabling,rarr,attributedqa,halogen} supply the primitives for \method's reference verifier.
(\romannumeral2) {Theoretical-claim verification.} LLM-based provers~\citep{deepseekproverv2,proveragent,hilbert,hermes} and agentic program verifiers~\citep{autorocq} combine informal reasoning with Lean and Coq checking.
\method\ incorporates this perspective: it densely extracts informal arguments and checks whether they match formal or executable content.
(\romannumeral3) {Experiment reproduction.} Beyond textual verification, reproducibility benchmarks~\citep{mlebench,mlagentbench,mlbench2} evaluate ML engineering on Kaggle- and repo-scale tasks, and others extend this to software, computational, scientific, and frontier-R\&D settings~\citep{swebench,corebench,scienceagentbench,rebench}, where top agents still fall below $40\%$ execution accuracy~\citep{researchcodebench,lmrbench,scireplicate}.
\method\ internalises the lesson that full reproduction is expensive and noisy: L3 turns the paper's own claims into a prioritised plan and executes it only with author approval, so compute goes where it is most informative.

\paragraph{Paper Peer Review}
A rapidly growing line casts LLMs as peer reviewers. Static prompting or fine-tuning yields a full review per paper~\citep{liang2023llmreferee,marg,reviewer2,deepreview,remor,lattereview}; multi-agent variants simulate the review cycle with role-specialised agents~\citep{agentreview,reviewmt,reviewgraph}; decomposition-based {TreeReview}~\citep{treereview} recursively asks sub-questions; and {CycleResearcher}~\citep{cyclereviewer} closes the loop via iterative preference optimization. Multimodal and multidisciplinary variants~\citep{mmreview,multimodalpeerreview} read figures and tables; pre-submission assistance has been proposed as an ethically cleaner deployment~\citep{foster2025openness}. At scale, the {Review Feedback Agent}~\citep{thakkar2025reviewfeedback} was deployed on 20k ICLR-2025 reviews. Parallel audits document persistent failure modes: prompt-injection susceptibility, sycophancy, and poor novelty calibration~\citep{llmrevalbias,llmreviewsurvey}.
These systems share a judge-oriented output contract, optimised to agree with held-out reviewer opinions rather than to serve the author, and they under-serve revision in three ways that \method\ inverts.
(i) Evidence grounding. Verdicts are rarely tied to a specific sentence, equation, or code line; \method\ emits (finding, evidence, suggestion) triples attached to concrete artefacts.
(ii) Claim-level auditability. Holistic judgements let individual unverifiable assertions pass silently; \method\ extracts claims and verifies each one.
(iii) Cost tiering. Review pipelines are binary, in that every check always runs or never runs; \method\ runs cheap screens by default, routes typed verifiers by the evidence they need, and reserves full reproduction for explicit author approval.

Existing methods are designed to ``\textit{judge}'' a finished manuscript, and are evaluated by how closely their verdicts agree with human reviewers. \method\ is designed to ``\textit{assist}'' a manuscript still in progress: it grounds every finding in verified evidence and pairs it with an actionable revision, so its success is measured by whether authors accept each finding and act on it, rather than by score agreement.

\section{Discussion}
\label{sec:conclusion}

We introduced \method, an agentic framework that shifts automated paper assessment from a {judge} to a {diagnostician}. Given a paper with its code and data, \method\ returns findings, each pairing an evidence with a location in the paper and a suggestion, through a pipeline that scales effort with the cost of verification: surface checks on the manuscript, typed verifiers that route each claim to the evidence that settles it, and selective reruns of experiments under a priority budget. Across a human study with junior researchers and 40 manuscripts spanning machine learning, the natural sciences, and the social sciences, \method\ produced more auditable feedback than human and other agentic reviewers, pairs critiques with suggestions by design, and complements dimensions that reading-only review often misses. Its reproduction stage reports where a claim could not be re-obtained under our protocol—including missing instructions, execution failures, and numeric disagreements—without treating venue or pass rate as a quality ranking.

\paragraph{Review for the paper, feedback for the author.}
A review is a judgment in service of a venue: accept or reject, with a justification that mostly explains the verdict. Feedback serves the author, and that is where \method\ sits. It automates the checkable part of feedback, such as cross-checking numbers against tables, aligning prose with code, and rerunning experiments, and leaves judgment alone: whether a question is worth asking and whether a result matters remain human calls. The aim is not to replace the human but to relocate their time and attention: once an agent has checked what can be checked, the hours an author spends on a draft can go to taste and direction. A natural extension is an interactive loop in which the author accepts or contests each finding and the affected checks re-run, so that machine coverage and human judgment compound.

\paragraph{Verifiable, executable environments for future papers.}
As agents begin to assist research and most human drafts carry some AI assistance, plausible-looking claims accumulate faster than anyone can check by hand, and assessment has to become verifiable and executable in turn. \method\ is built around this principle. Every feedback can be audited: its anchor points to the exact sentence, equation, line of code, or reference it critiques, so a fabricated critique fails at its own anchor and the author can dismiss it. 
If diagnosis of this kind becomes routine, a claim backed by runnable code and traceable numbers becomes cheap to check, while a claim without them stays expensive to trust.

\paragraph{Future work.}
Two directions stand out. (i) {Full paper-to-code reproduction.} \method's strongest evidence depends on a runnable codebase. When authors release none, an agent could synthesise a reference implementation directly from the paper itself, as PaperBench~\cite{paperbench} does, though this setting will yield lower reproduction rates than the current system. (ii) {Large-scale advisor-level studies.} Advisor comments are rarely recorded systematically and mostly exist as sparse, informal remarks. Collecting such feedback at scale, and grounding it against external reviewer and AI feedback, would let us capture the insight that only experienced advisors provide. We release \method\ together with a demo interface, in the hope of supporting more rigorous and reproducible research in the community.

\appendix
\newpage
\section{\method~Interface}
\label{sec:interface}

\begin{figure*}[!h]
    \centering
    \includegraphics[width=\linewidth]{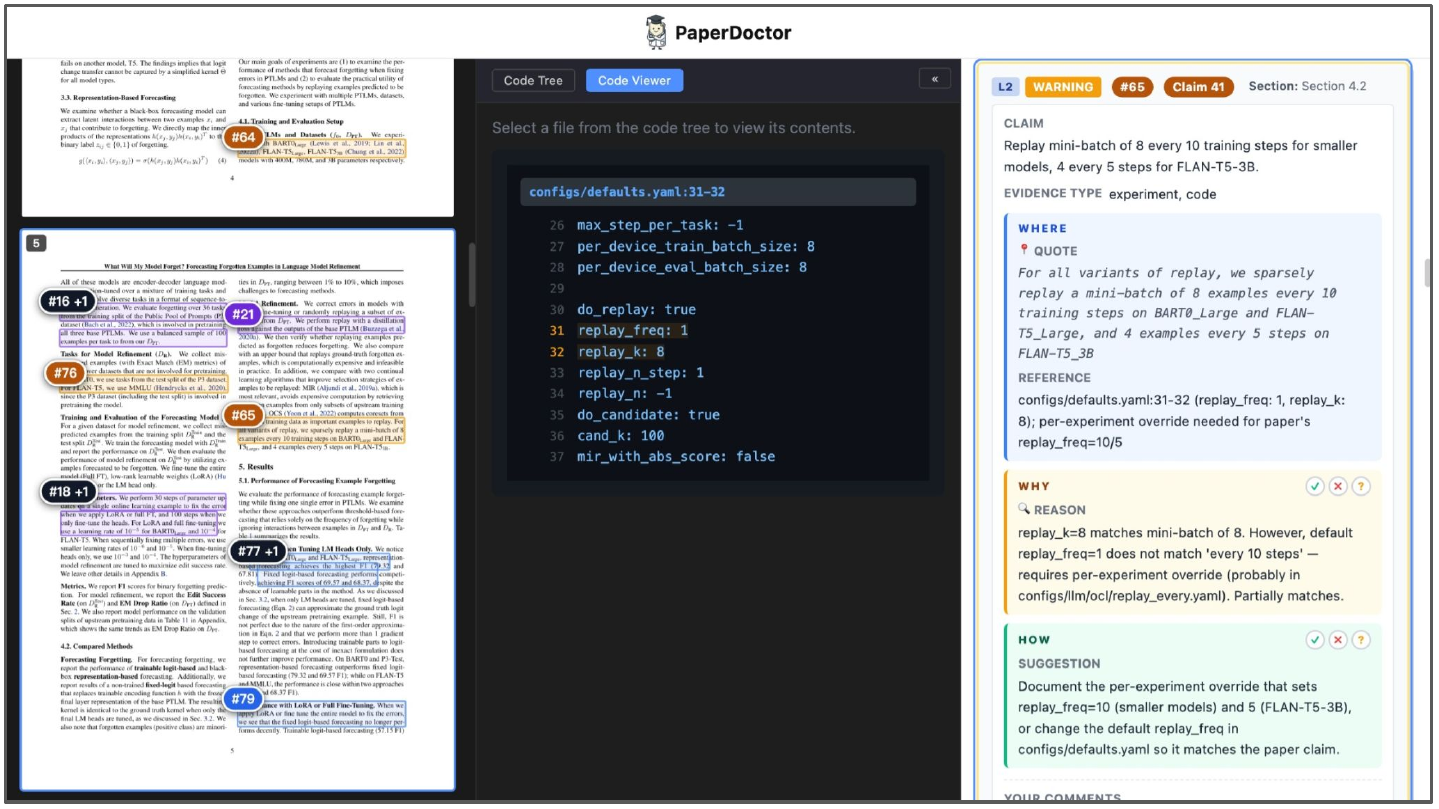}
\caption{\textbf{The \method\ Diagnosis Interface.} Authors upload their paper and code, and \method\ returns the holistic report. \textbf{Left:} the paper, with each finding overlaid as a numbered, color-coded bubble anchored to the exact span it critiques (Where). \textbf{Middle:} the code viewer, so findings about implementation can be audited against the source. \textbf{Right:} the finding card for the selected highlight, showing the observation (Why) and a concrete suggestion for revision (How). 
In the example shown, \method\ identifies an implementation-level mismatch.
: the paper claims a replay mini-batch of 8 every 10 training steps, but the default config (\texttt{configs/defaults.yaml:31-32}) sets \texttt{replay\_freq=1}.
}
\label{fig:demo}
\end{figure*}

\section{Case Studies}
\label{sec:case-study}

In this section, we walk through representative cases drawn from our 40-paper testbed. Each case follows the reason (why), quote evidence (where) and suggestion (how).

\newpage

\noindent\textbf{L1 Citation Check: a fabricated future date.}\par
\vspace{2pt}\noindent\includegraphics[width=0.9\columnwidth]{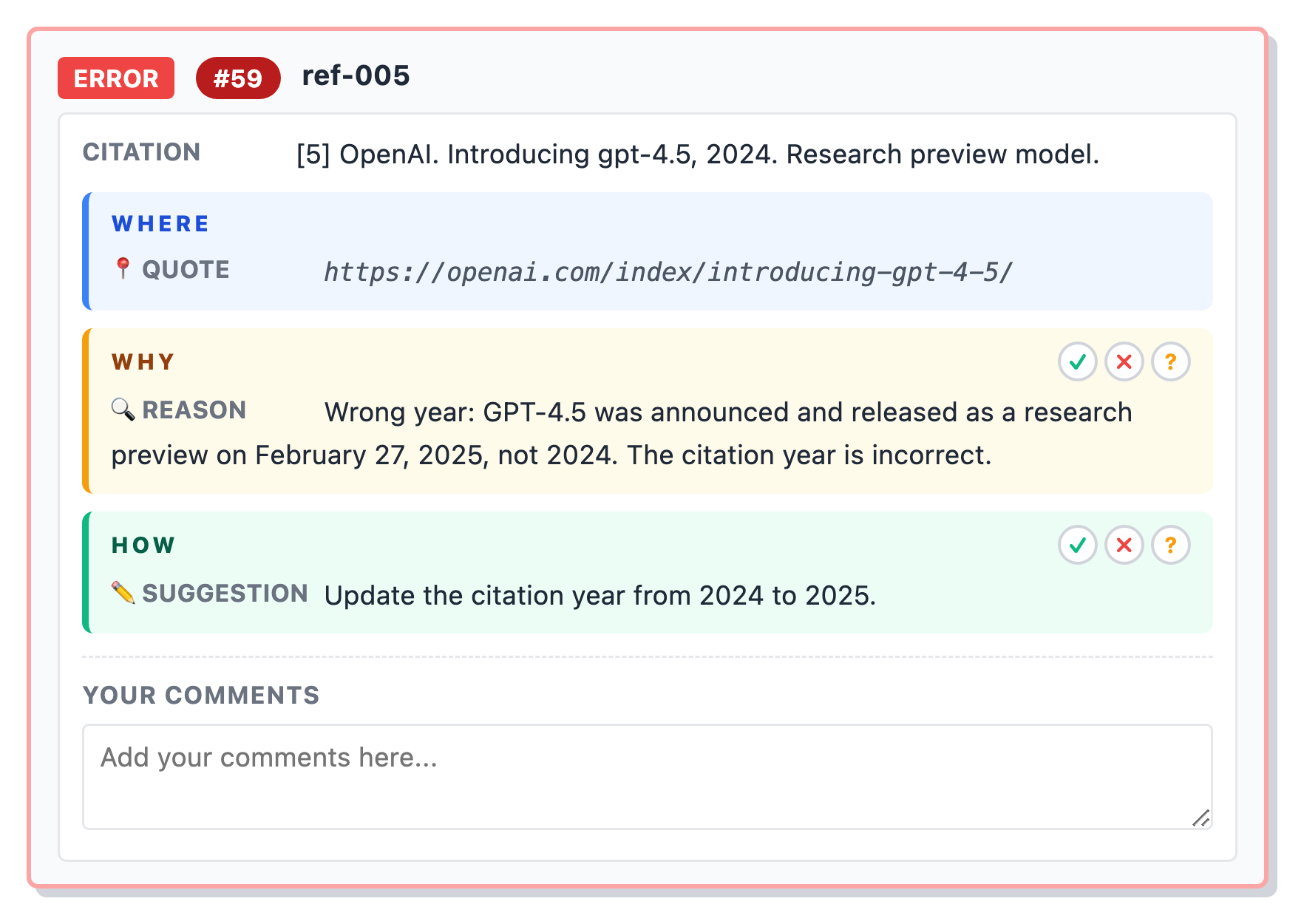}\par
\vspace{2pt}\noindent In \textit{Reasoning Models Outperform Standard Language Models in De Novo Protein Design} (Agents4Science), the bibliography contains the entry ``[5] OpenAI. Introducing GPT-4.5, 2024. Research preview model.'' \method's citation verifier issues a web search for this reference and finds that \textit{GPT-4.5} was announced and released on February~27, 2025, not in 2024. Although the manuscript itself appeared in late 2024, the cited release date is therefore chronologically impossible. Standard citation checkers accept any correctly-formatted entry; only a search-grounded verifier flags this kind of temporally inconsistent metadata, which is a recurring failure mode of AI-assisted drafts that synthesise plausible-sounding venues and years without grounding them in real release notes.

\newpage
\noindent\textbf{L2 Code Verification: paper claims an encoder its code never instantiates.}\par
\vspace{2pt}\noindent\includegraphics[width=0.9\columnwidth]{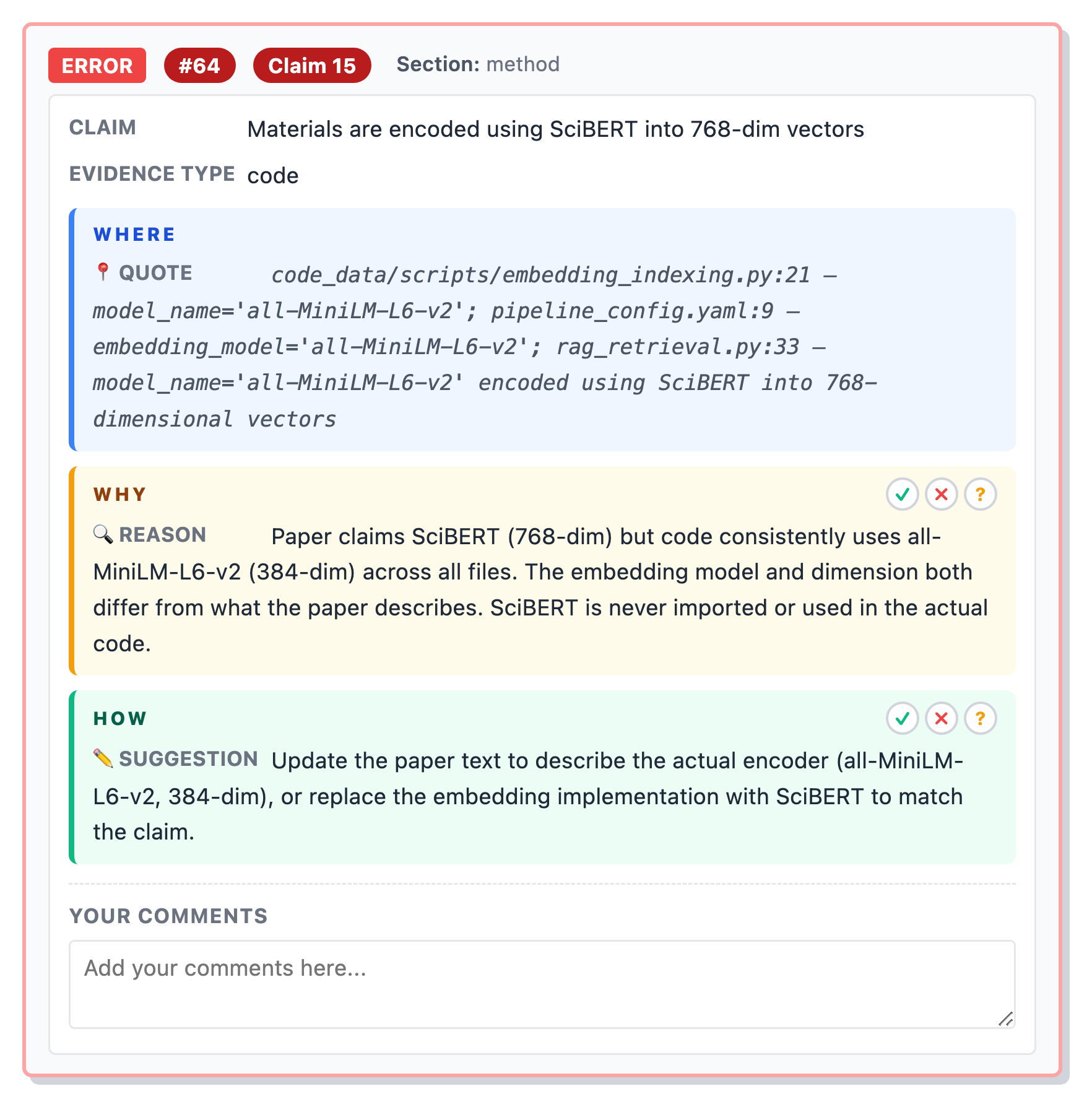}\par
\vspace{2pt}\noindent \textit{LLM-Driven Discovery of High-Entropy Catalysts via Retrieval-Augmented Generation} (a second Agents4Science paper) states in its Methods section that materials are ``encoded using \textit{SciBERT} into 768-dimensional vectors''. \method's code verifier dispatches the embedding-model claim to the released repository and finds that all three relevant files, \texttt{code\_data/\allowbreak scripts/\allowbreak embedding\_\allowbreak indexing.py} (line~21), \texttt{code\_data/\allowbreak scripts/\allowbreak rag\_\allowbreak retrieval.py} (line~33), and \texttt{pipeline\_\allowbreak config.yaml} (line~9), set \texttt{model\_name='\allowbreak all-MiniLM-L6-v2'}. \textit{SciBERT} is never imported anywhere in the codebase, and the produced vectors are 384-dimensional rather than 768. The discrepancy spans both the model identity and its dimensionality. This is a paradigmatic case of a mismatch that is invisible to reading-only review: the paper text reads cleanly, but the implementation differs substantively from what is described.

\newpage
\noindent\textbf{L2 Theory Verification: a bound stated without definitions.}\par
\vspace{2pt}\noindent\includegraphics[width=0.9\columnwidth]{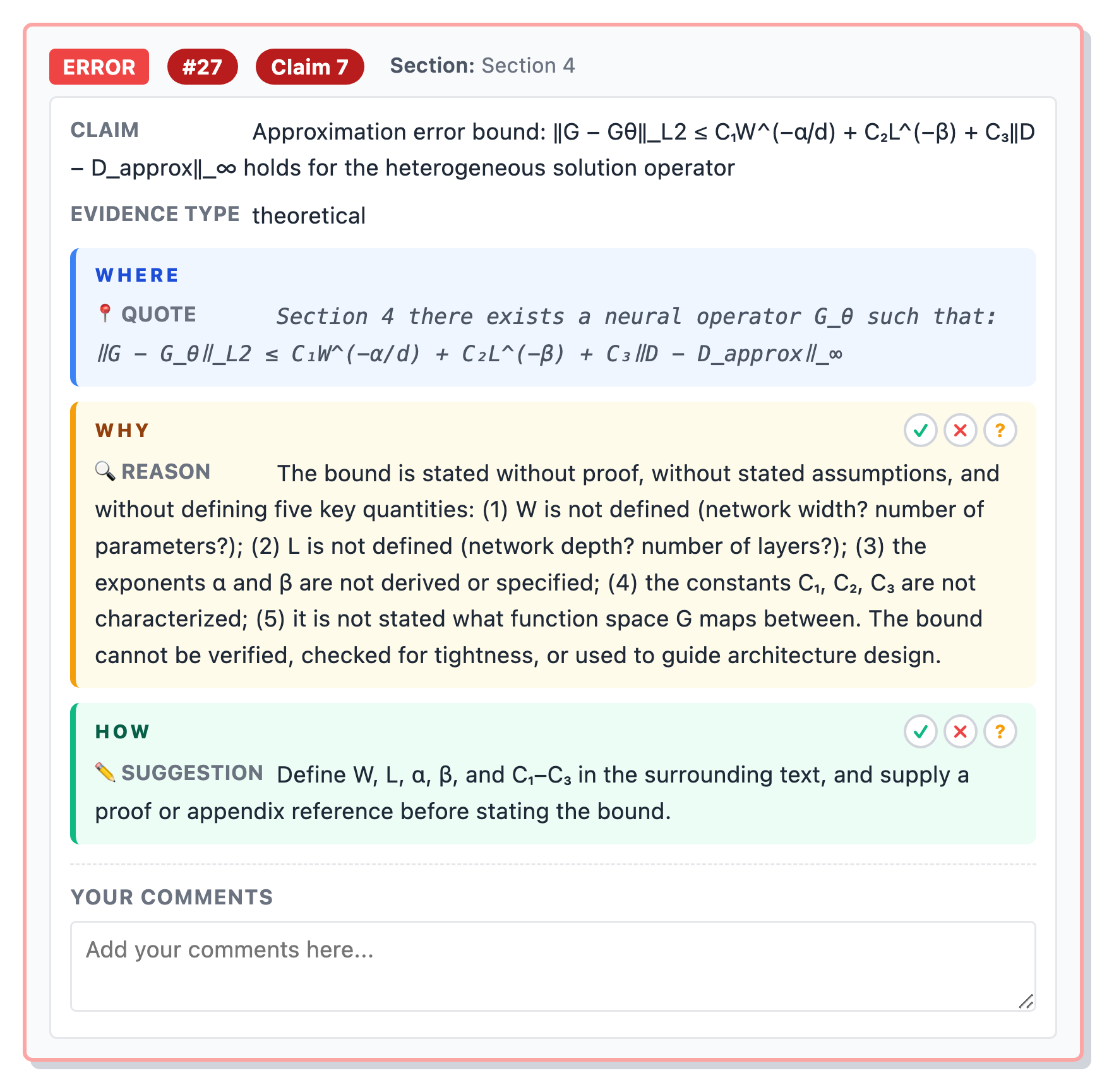}\par
\vspace{2pt}\noindent In \textit{Neural Reaction-Diffusion Operators for Spatially Heterogeneous Tumor Modeling} (Agents4Science), Section~4 states the approximation bound $\|G - G_\theta\|_{L^2} \leq C_1 W^{-\alpha/d} + C_2 L^{-\beta} + C_3 \|D - D_{\text{approx}}\|_\infty$ as if it were established. \method's step-by-step re-derivation finds that none of the symbols composing the bound are introduced anywhere in the paper: $W$ (presumably network width), $L$ (presumably depth), the exponents $\alpha$ and $\beta$, and the constants $C_1$, $C_2$, $C_3$ all appear without definition, and no proof or pointer to an appendix is given. Without these, the bound cannot be checked for tightness or used to guide architecture design. On a first pass, such a gap looks like a finished theorem; only step-by-step re-derivation surfaces it.

\newpage
\noindent\textbf{L2 Literature Check: a novelty claim refuted by the cited works.}\par
\vspace{2pt}\noindent\includegraphics[width=0.9\columnwidth]{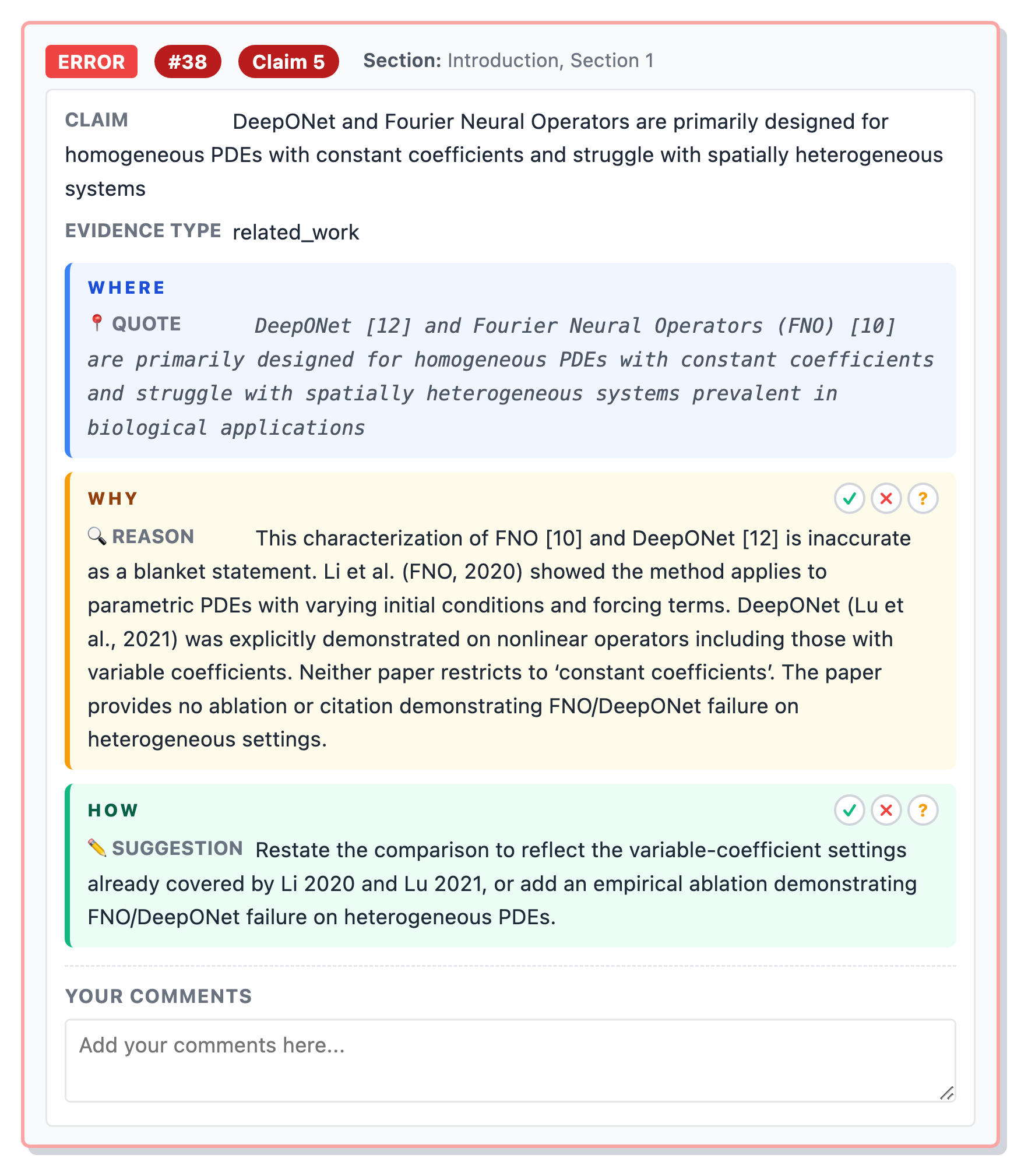}\par
\vspace{2pt}\noindent The same paper claims that ``\textit{DeepONet}~[12] and \textit{Fourier Neural Operators} (FNO)~[10] are primarily designed for homogeneous PDEs with constant coefficients and struggle with spatially heterogeneous systems prevalent in biological applications.'' \method~dispatches this novelty assertion as a web search query, and the original method papers refute the framing directly: Li et~al.~(FNO, 2020,~[10]) explicitly demonstrates the method on parametric PDEs with varying initial conditions and forcing terms, and Lu et~al.~(DeepONet, 2021,~[12]) demonstrates nonlinear operators with variable coefficients. Neither prior work is restricted to the homogeneous setting the paper attributes to it. This pattern, where an inflated novelty claim is contradicted by the very references the paper cites, is exactly what a search-grounded literature check is built to catch.

\newpage
\noindent\textbf{L3 Reproduction: a ``factor of 2'' that is actually $1.14\times$.}\par
\vspace{2pt}\noindent\includegraphics[width=0.9\columnwidth]{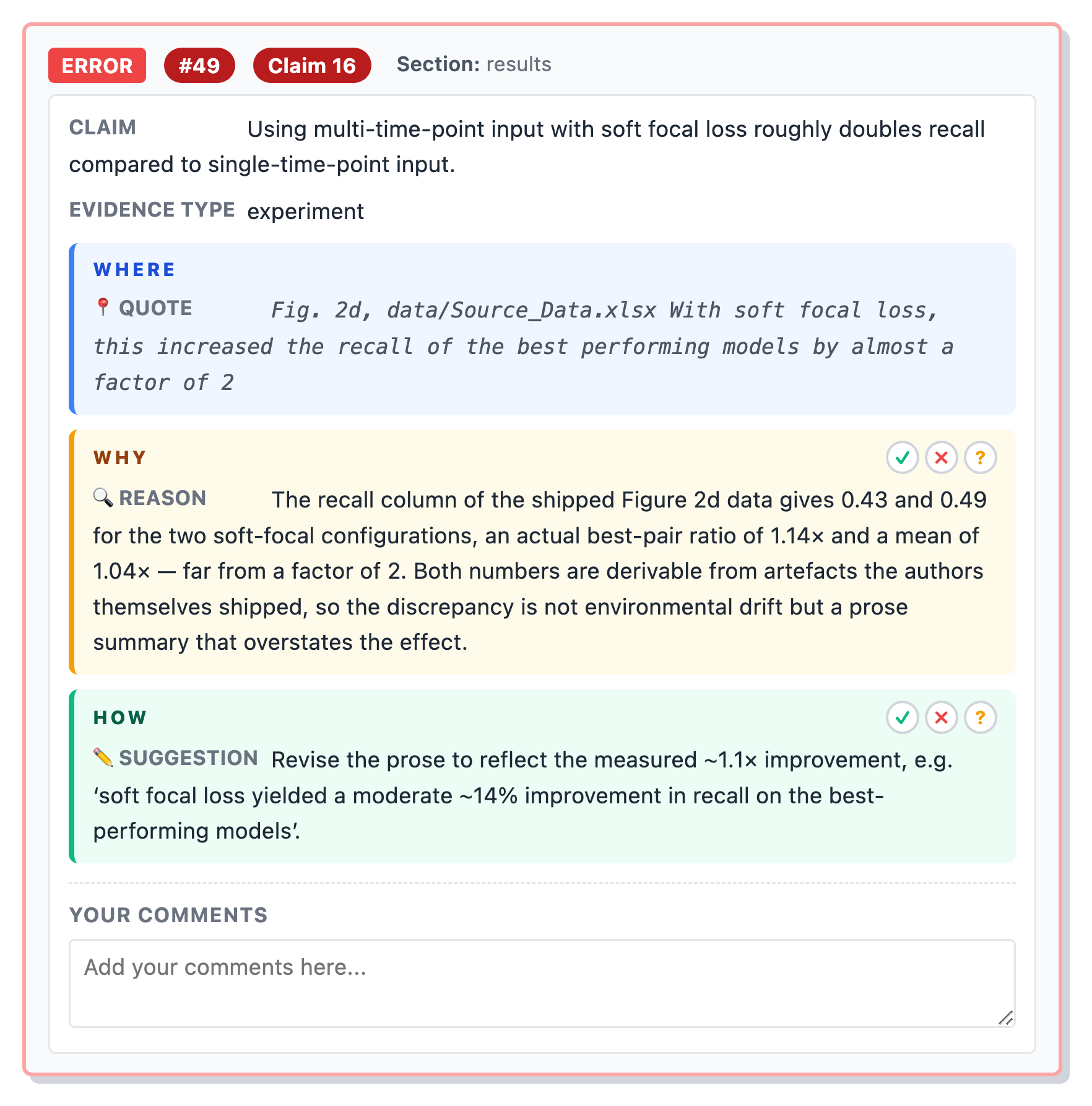}\par
\vspace{2pt}\noindent In \textit{Smart hybrid microscopy for cell-friendly detection of rare events}, a Nature Communications paper on mitochondrial imaging, the Results section claims that ``soft focal loss \dots~increased the recall of the best performing models by almost a factor of $2$''. The L3 stage rebuilds the analysis from the source data shipped with the paper and reads the recall column for Figure~2d directly. The two soft-focal configurations have contact-task recalls of $0.43$ and $0.49$, giving an actual best-pair ratio of $1.14\times$ and a mean ratio of $1.04\times$, far from the claimed factor of two. Both the claimed and reproduced numbers are derivable from artefacts the authors themselves shipped, so the discrepancy is not a matter of environmental drift but of how a quantitative claim was summarised in the text. The aggregate reproduction analysis (Sec.~\ref{sec:exp:repro}) shows that even after execution is feasible, reproduced quantities can still diverge from published claims; the case here shows that peer-reviewed venues are not immune, and that reading the paper alone cannot easily catch it.


\bibliographystyle{plainnat}
\bibliography{main}

@article{aiscientist,
  title={The ai scientist: Towards fully automated open-ended scientific discovery},
  author={Lu, Chris and Lu, Cong and Lange, Robert Tjarko and Foerster, Jakob and Clune, Jeff and Ha, David},
  journal={arXiv preprint arXiv:2408.06292},
  year={2024}
}

@article{aiscientistv2,
  title={The ai scientist-v2: Workshop-level automated scientific discovery via agentic tree search},
  author={Yamada, Yutaro and Lange, Robert Tjarko and Lu, Cong and Hu, Shengran and Lu, Chris and Foerster, Jakob and Clune, Jeff and Ha, David},
  journal={arXiv preprint arXiv:2504.08066},
  year={2025}
}

@article{agent4science,
  title={Exploring the use of AI authors and reviewers at Agents4Science},
  author={Bianchi, Federico and Queen, Owen and Thakkar, Nitya and Sun, Eric and Zou, James},
  journal={Nature Biotechnology},
  volume={44},
  number={1},
  pages={11--14},
  year={2026},
  publisher={Nature Publishing Group US New York}
}

@article{paperbench,
  title={PaperBench: Evaluating AI's Ability to Replicate AI Research},
  author={Starace, Giulio and Jaffe, Oliver and Sherburn, Dane and Aung, James and Chan, Jun Shern and Maksin, Leon and Dias, Rachel and Mays, Evan and Kinsella, Benjamin and Thompson, Wyatt and others},
  journal={arXiv preprint arXiv:2504.01848},
  year={2025}
}

@inproceedings{agentreview,
  title={AgentReview: Exploring Peer Review Dynamics with LLM Agents},
 author={Jin, Yiqiao and Zhao, Qinlin and Wang, Yiyang and Chen, Hao and Zhu, Kaijie and Xiao, Yijia and Wang, Jindong},
  booktitle={Proceedings of the 2024 Conference on Empirical Methods in Natural Language Processing},
  pages={1208--1226},
  year={2024}
}

@article{llmreviewsurvey,
  title={Large language models for automated scholarly paper review: A survey},
  author={Zhuang, Zhenzhen and Chen, Jiandong and Xu, Hongfeng and Jiang, Yuwen and Lin, Jialiang},
  journal={Information Fusion},
  volume={124},
  pages={103332},
  year={2025},
  publisher={Elsevier}
}

@article{agentlaboratory,
  title={Agent Laboratory: Using {LLM} Agents as Research Assistants},
  author={Schmidgall, Samuel and Su, Yusheng and Wang, Ze and Sun, Ximeng and Wu, Jialian and Yu, Xiaodong and Liu, Jiang and Liu, Zicheng and Barsoum, Emad},
  journal={arXiv preprint arXiv:2501.04227},
  year={2025}
}

@article{airesearcher,
  title={AI-Researcher: Autonomous Scientific Innovation},
  author={Tang, Jiabin and Xia, Lianghao and Li, Zhonghang and Huang, Chao},
  journal={arXiv preprint arXiv:2505.18705},
  year={2025}
}

@article{researchcodebench,
  title={ResearchCodeBench: Benchmarking {LLM}s on Implementing Novel Machine Learning Research Code},
  author={Hua, Tianyu and Hua, Harper and Xiang, Violet and Klieger, Benjamin and Truong, Sang and Liang, Weixin and Sun, Fan-Yun and Haber, Nick},
  journal={Advances in Neural Information Processing Systems},
  volume={38},
  year={2026}
}

@article{lmrbench,
  title={Lmr-bench: Evaluating llm agent’s ability on reproducing language modeling research, 2025},
  author={Yan, Shuo and Li, Ruochen and Luo, Ziming and Wang, Zimu and Li, Daoyang and Jing, Liqiang and He, Kaiyu and Wu, Peilin and Michalopoulos, George and Zhang, Yue and others},
  journal={URL https://arxiv. org/abs/2506.17335}
}

@article{scireplicate,
  title={{SciReplicate-Bench}: Benchmarking {LLM}s in Agent-driven Algorithmic Reproduction from Research Papers},
  author={Xiang, Yanzheng and Yan, Hanqi and Ouyang, Shuyin and Gui, Lin and He, Yulan},
  journal={arXiv preprint arXiv:2504.00255},
  year={2025}
}

@article{citeaudit,
  title={CiteAudit: You Cited It, But Did You Read It? A Benchmark for Verifying Scientific References in the LLM Era},
  author={Yuan, Zhengqing and Shi, Kaiwen and Zhang, Zheyuan and Sun, Lichao and Chawla, Nitesh V and Ye, Yanfang},
  journal={arXiv preprint arXiv:2602.23452},
  year={2026}
}

@article{mmreview,
  title={{MMReview}: A Multidisciplinary and Multimodal Benchmark for {LLM}-Based Peer Review Automation},
  author={Gao, Xian and Ruan, Jiacheng and Zhang, Zongyun and Gao, Jingsheng and Liu, Ting and Fu, Yuzhuo},
  journal={arXiv preprint arXiv:2508.14146},
  year={2025}
}

@inproceedings{marg,
  title={{MARG}: Multi-Agent Review Generation for Scientific Papers},
  author={D'Arcy, Mike and Hope, Tom and Birnbaum, Larry and Downey, Doug},
  booktitle={arXiv preprint arXiv:2401.04259},
  year={2024}
}

@article{liang2023llmreferee,
  title={Can large language models provide useful feedback on research papers? A large-scale empirical analysis},
  author={Liang, Weixin and Zhang, Yuhui and Cao, Hancheng and Wang, Binglu and Ding, Daisy Yi and Yang, Xinyu and Vodrahalli, Kailas and He, Siyu and Smith, Daniel Scott and Yin, Yian and others},
  journal={NEJM AI},
  volume={1},
  number={8},
  pages={AIoa2400196},
  year={2024},
  publisher={Massachusetts Medical Society}
}

@article{reviewer2,
  title={Reviewer2: Optimizing review generation through prompt generation},
  author={Gao, Zhaolin and Brantley, Kiant{\'e} and Joachims, Thorsten},
  journal={arXiv preprint arXiv:2402.10886},
  year={2024}
}

@inproceedings{deepreview,
  title={Deepreview: Improving llm-based paper review with human-like deep thinking process},
  author={Zhu, Minjun and Weng, Yixuan and Yang, Linyi and Zhang, Yue},
  booktitle={Proceedings of the 63rd Annual Meeting of the Association for Computational Linguistics (Volume 1: Long Papers)},
  pages={29330--29355},
  year={2025}
}

@inproceedings{treereview,
  title={TreeReview: A dynamic tree of questions framework for deep and efficient LLM-based scientific peer review},
  author={Chang, Yuan and Li, Ziyue and Zhang, Hengyuan and Kong, Yuanbo and Wu, Yanru and So, Hayden Kwok-Hay and Guo, Zhijiang and Zhu, Liya and Wong, Ngai},
  booktitle={Proceedings of the 2025 Conference on Empirical Methods in Natural Language Processing},
  pages={15662--15693},
  year={2025}
}

@article{cycleresearcher,
  title={Cycleresearcher: Improving automated research via automated review},
  author={Weng, Yixuan and Zhu, Minjun and Bao, Guangsheng and Zhang, Hongbo and Wang, Jindong and Zhang, Yue and Yang, Linyi},
  journal={arXiv preprint arXiv:2411.00816},
  year={2024}
}

@article{remor,
  title={Remor: Automated peer review generation with llm reasoning and multi-objective reinforcement learning},
  author={Taechoyotin, Pawin and Acuna, Daniel},
  journal={arXiv preprint arXiv:2505.11718},
  year={2025}
}

@article{lattereview,
  title={LatteReview: a multi-agent framework for systematic review automation using large language models},
  author={Rouzrokh, Pouria and Khosravi, Bardia and Rouzrokh, Parsa and Shariatnia, Moein},
  journal={arXiv preprint arXiv:2501.05468},
  year={2025}
}

@article{reviewmt,
  title={Peer review as a multi-turn and long-context dialogue with role-based interactions},
  author={Tan, Cheng and Lyu, Dongxin and Li, Siyuan and Gao, Zhangyang and Wei, Jingxuan and Ma, Siqi and Liu, Zicheng and Li, Stan Z},
  journal={arXiv preprint arXiv:2406.05688},
  year={2024}
}

@inproceedings{reviewgraph,
  title={Automatic paper reviewing with heterogeneous graph reasoning over llm-simulated reviewer-author debates},
  author={Li, Shuaimin and Fan, Liyang and Lin, Yufang and Li, Zeyang and Wei, Xian and Ni, Shiwen and Alinejad-Rokny, Hamid and Yang, Min},
  booktitle={Proceedings of the AAAI Conference on Artificial Intelligence},
  volume={40},
  number={37},
  pages={31717--31725},
  year={2026}
}

@article{multimodalpeerreview,
  title={Multimodal Peer Review Simulation with Actionable To-Do Recommendations for Community-Aware Manuscript Revisions},
  author={Hong, Mengze and Jiang, Di and Zhao, Weiwei and Li, Yawen and Wang, Yihang and Luo, Xinyuan and Sun, Yanjie and Zhang, Chen Jason},
  journal={arXiv preprint arXiv:2511.10902},
  year={2025}
}

@article{thakkar2025reviewfeedback,
  title={Can LLM feedback enhance review quality? A randomized study of 20k reviews at ICLR 2025},
  author={Thakkar, Nitya and Yuksekgonul, Mert and Silberg, Jake and Garg, Animesh and Peng, Nanyun and Sha, Fei and Yu, Rose and Vondrick, Carl and Zou, James},
  journal={arXiv preprint arXiv:2504.09737},
  year={2025}
}

@article{llmrevalbias,
  title={LLM-REVal: Can We Trust LLM Reviewers Yet?},
  author={Li, Rui and Gu, Jia-Chen and Kung, Po-Nien and Xia, Heming and Kong, Xiangwen and Sui, Zhifang and Peng, Nanyun and others},
  journal={arXiv preprint arXiv:2510.12367},
  year={2025}
}

@article{foster2025openness,
  title={Openness in AI and downstream governance: A global value chain approach},
  author={Foster, Christopher},
  journal={arXiv preprint arXiv:2509.10220},
  year={2025}
}

@article{cyclereviewer,
  title={Cycleresearcher: Improving automated research via automated review},
  author={Weng, Yixuan and Zhu, Minjun and Bao, Guangsheng and Zhang, Hongbo and Wang, Jindong and Zhang, Yue and Yang, Linyi},
  journal={arXiv preprint arXiv:2411.00816},
  year={2024}
}

@article{ghostcite,
  title={GhostCite: A Large-Scale Analysis of Citation Validity in the Age of Large Language Models},
  author={Xu, Zuyao and Qiu, Yuqi and Sun, Lu and Miao, FaSheng and Wu, Fubin and Wang, Xinyi and Li, Xiang and Lu, Haozhe and Zhang, ZhengZe and Hu, Yuxin and others},
  journal={arXiv preprint arXiv:2602.06718},
  year={2026}
}

@article{citehallucinationstudy,
  title={How LLMs Cite and Why It Matters: A Cross-Model Audit of Reference Fabrication in AI-Assisted Academic Writing and Methods to Detect Phantom Citations},
  author={Naser, MZ},
  journal={arXiv preprint arXiv:2603.03299},
  year={2026}
}

@article{wu2025automated,
  title={An automated framework for assessing how well LLMs cite relevant medical references},
  author={Wu, Kevin and Wu, Eric and Wei, Kevin and Zhang, Angela and Casasola, Allison and Nguyen, Teresa and Riantawan, Sith and Shi, Patricia and Ho, Daniel and Zou, James},
  journal={Nature Communications},
  volume={16},
  number={1},
  pages={3615},
  year={2025},
  publisher={Nature Publishing Group UK London}
}

@inproceedings{gao2023enabling,
  title={Enabling large language models to generate text with citations},
  author={Gao, Tianyu and Yen, Howard and Yu, Jiatong and Chen, Danqi},
  booktitle={Proceedings of the 2023 Conference on Empirical Methods in Natural Language Processing},
  pages={6465--6488},
  year={2023}
}

@inproceedings{rarr,
  title={Rarr: Researching and revising what language models say, using language models},
  author={Gao, Luyu and Dai, Zhuyun and Pasupat, Panupong and Chen, Anthony and Chaganty, Arun Tejasvi and Fan, Yicheng and Zhao, Vincent and Lao, Ni and Lee, Hongrae and Juan, Da-Cheng and others},
  booktitle={Proceedings of the 61st Annual Meeting of the Association for Computational Linguistics (Volume 1: Long Papers)},
  pages={16477--16508},
  year={2023}
}

@article{attributedqa,
  title={Attributed question answering: Evaluation and modeling for attributed large language models},
  author={Bohnet, Bernd and Tran, Vinh Q and Verga, Pat and Aharoni, Roee and Andor, Daniel and Soares, Livio Baldini and Ciaramita, Massimiliano and Eisenstein, Jacob and Ganchev, Kuzman and Herzig, Jonathan and others},
  journal={arXiv preprint arXiv:2212.08037},
  year={2022}
}

@inproceedings{halogen,
  title={Halogen: Fantastic llm hallucinations and where to find them},
  author={Ravichander, Abhilasha and Ghela, Shrusti and Wadden, David and Choi, Yejin},
  booktitle={Proceedings of the 63rd Annual Meeting of the Association for Computational Linguistics (Volume 1: Long Papers)},
  pages={1402--1425},
  year={2025}
}

@article{deepseekproverv2,
  title={Deepseek-prover-v2: Advancing formal mathematical reasoning via reinforcement learning for subgoal decomposition},
  author={Ren, ZZ and Shao, Zhihong and Song, Junxiao and Xin, Huajian and Wang, Haocheng and Zhao, Wanjia and Zhang, Liyue and Fu, Zhe and Zhu, Qihao and Yang, Dejian and others},
  journal={arXiv preprint arXiv:2504.21801},
  year={2025}
}

@article{proveragent,
  title={Prover agent: An agent-based framework for formal mathematical proofs},
  author={Baba, Kaito and Liu, Chaoran and Kurita, Shuhei and Sannai, Akiyoshi},
  journal={arXiv preprint arXiv:2506.19923},
  year={2025}
}

@article{hilbert,
  title={Hilbert: Recursively building formal proofs with informal reasoning},
  author={Varambally, Sumanth and Voice, Thomas and Sun, Yanchao and Chen, Zhifeng and Yu, Rose and Ye, Ke},
  journal={arXiv preprint arXiv:2509.22819},
  year={2025}
}

@article{hermes,
  title={HERMES: Towards Efficient and Verifiable Mathematical Reasoning in LLMs},
  author={Ospanov, Azim and Feng, Zijin and Sun, Jiacheng and Bai, Haoli and Shen, Xin and Farnia, Farzan},
  journal={arXiv preprint arXiv:2511.18760},
  year={2025}
}

@article{autorocq,
  title={Agentic Program Verification},
  author={Tu, Haoxin and Zhao, Huan and Song, Yahui and Zafar, Mehtab and Meng, Ruijie and Roychoudhury, Abhik},
  journal={arXiv preprint arXiv:2511.17330},
  year={2025}
}

@article{mlebench,
  title={Mle-bench: Evaluating machine learning agents on machine learning engineering},
  author={Chan, Jun Shern and Chowdhury, Neil and Jaffe, Oliver and Aung, James and Sherburn, Dane and Mays, Evan and Starace, Giulio and Liu, Kevin and Maksin, Leon and Patwardhan, Tejal and others},
  journal={arXiv preprint arXiv:2410.07095},
  year={2024}
}

@article{mlagentbench,
  title={Mlagentbench: Evaluating language agents on machine learning experimentation},
  author={Huang, Qian and Vora, Jian and Liang, Percy and Leskovec, Jure},
  journal={arXiv preprint arXiv:2310.03302},
  year={2023}
}

@article{mlbench2,
  title={Ml-bench: Evaluating large language models and agents for machine learning tasks on repository-level code},
  author={Tang, Xiangru and Liu, Yuliang and Cai, Zefan and Shao, Yanjun and Lu, Junjie and Zhang, Yichi and Deng, Zexuan and Hu, Helan and An, Kaikai and Huang, Ruijun and others},
  journal={arXiv preprint arXiv:2311.09835},
  year={2023}
}

@article{scienceagentbench,
  title={Scienceagentbench: Toward rigorous assessment of language agents for data-driven scientific discovery},
  author={Chen, Ziru and Chen, Shijie and Ning, Yuting and Zhang, Qianheng and Wang, Boshi and Yu, Botao and Li, Yifei and Liao, Zeyi and Wei, Chen and Lu, Zitong and others},
  journal={arXiv preprint arXiv:2410.05080},
  year={2024}
}

@article{swebench,
  title={Swe-bench: Can language models resolve real-world github issues?},
  author={Jimenez, Carlos E and Yang, John and Wettig, Alexander and Yao, Shunyu and Pei, Kexin and Press, Ofir and Narasimhan, Karthik},
  journal={arXiv preprint arXiv:2310.06770},
  year={2023}
}

@article{corebench,
  title={Core-bench: Fostering the credibility of published research through a computational reproducibility agent benchmark},
  author={Siegel, Zachary S and Kapoor, Sayash and Nagdir, Nitya and Stroebl, Benedikt and Narayanan, Arvind},
  journal={arXiv preprint arXiv:2409.11363},
  year={2024}
}

@article{rebench,
  title={Re-bench: Evaluating frontier ai r\&d capabilities of language model agents against human experts},
  author={Wijk, Hjalmar and Lin, Tao and Becker, Joel and Jawhar, Sami and Parikh, Neev and Broadley, Thomas and Chan, Lawrence and Chen, Michael and Clymer, Josh and Dhyani, Jai and others},
  journal={arXiv preprint arXiv:2411.15114},
  year={2024}
}

@article{reviewfeedback,
  title={A large-scale randomized study of large language model feedback in peer review},
  author={Thakkar, Nitya and Yuksekgonul, Mert and Silberg, Jake and Garg, Animesh and Peng, Nanyun and Sha, Fei and Yu, Rose and Vondrick, Carl and Zou, James},
  journal={Nature Machine Intelligence},
  pages={1--11},
  year={2026},
  publisher={Nature Publishing Group UK London}
}

@article{can2011model,
  title={A model for doctoral students’ perceptions and attitudes toward written feedback for academic writing},
  author={Can, Gulfidan and Walker, Andrew},
  journal={Research in Higher Education},
  volume={52},
  number={5},
  pages={508--536},
  year={2011},
  publisher={Springer}
}

@article{steiss2024comparing,
  title={Comparing the quality of human and ChatGPT feedback of students’ writing},
  author={Steiss, Jacob and Tate, Tamara and Graham, Steve and Cruz, Jazmin and Hebert, Michael and Wang, Jiali and Moon, Youngsun and Tseng, Waverly and Warschauer, Mark and Olson, Carol Booth},
  journal={Learning and Instruction},
  volume={91},
  pages={101894},
  year={2024},
  publisher={Elsevier}
}

@article{autoreproduce,
  title={Autoreproduce: Automatic ai experiment reproduction with paper lineage},
  author={Zhao, Xuanle and Sang, Zilin and Li, Yuxuan and Shi, Qi and Zhao, Weilun and Wang, Shuo and Zhang, Duzhen and Han, Xu and Liu, Zhiyuan and Sun, Maosong},
  journal={arXiv preprint arXiv:2505.20662},
  year={2025}
}
\end{document}